\documentclass[11pt,fancychapters]{report}
\usepackage[a4paper, total={6in, 8in}]{geometry}
\usepackage{listings}
\usepackage{color}
\usepackage{setspace}
\usepackage{acro}
\usepackage{amsmath}
\usepackage{amsthm}
\usepackage{graphicx}
\usepackage{geometry}
\usepackage{subcaption}
\usepackage{cancel}
\usepackage{tikz}
\usepackage{color}
\usepackage[frozencache,cachedir=.]{minted}
\usepackage{caption}
\usepackage{comment}
\usepackage{glossaries}
\usepackage{authblk}
\usepackage[backend=bibtex]{biblatex}
\usepackage{doi}

\usepackage{pythontex}

\newcounter{example}[section]
\newenvironment{example}[1][]{\refstepcounter{example}\par\medskip
   \noindent \textbf{Example~\theexample. #1} \rmfamily}{\medskip}

\newtheorem{theorem}{Theorem}[section]   
\newtheorem{exercise}[theorem]{Exercise}
\newtheorem{solution}[theorem]{Solution}
\theoremstyle{definition}
\newtheorem{definition}{Definition}[section]
\newtheorem{lemma}[theorem]{Lemma}

\usetikzlibrary{calc,trees,positioning,arrows,chains,shapes.geometric,%
    decorations.pathreplacing,decorations.pathmorphing,shapes,%
    matrix,shapes.symbols}
\DeclareAcronym{etf}{
	short = ETF,
    long = Exchange-Traded Fund,
    tag = abbrev
}

\DeclareAcronym{aum}{
	short = AUM,
    long = Assets Under Management,
    tag = abbrev
}

\DeclareAcronym{sr}{
	short = SR,
    long = Sharpe Ratio,
    tag = abbrev
}

\DeclareAcronym{nyse}{
	short = NYSE,
    long = New York Stock Exchange,
    tag = abbrev
}

\DeclareAcronym{hft}{
	short = HFT,
    long = High Frequency Trading,
    tag = abbrev
}

\DeclareAcronym{pv}{
	short = PV,
    long = present value,
    tag = abbrev
}

\DeclareAcronym{fv}{
	short = FV,
    long = future value,
    tag = abbrev
}

\DeclareAcronym{ir}{
	short = IR,
    long = interest rate,
    tag = abbrev
}

\DeclareAcronym{capm}{
	short = CAPM,
    long = Capital Assets Pricing Model,
    tag = abbrev
}

\DeclareAcronym{apt}{
	short = APT,
    long = Arbitrage Pricing Theory,
    tag = abbrev
}

\DeclareAcronym{sma}{
	short = SMA,
    long = simple moving average,
    tag = abbrev
}

\DeclareAcronym{mvo}{
	short = MVO,
    long = mean variance optimization,
    tag = abbrev
}

\DeclareAcronym{rmse}{
	short = RMSE,
    long = root mean square error,
    tag = abbrev
}

\DeclareAcronym{kNN}{
	short = kNN,
    long = k-nearest neighbor,
    tag = abbrev
}
\hypersetup{
    colorlinks,
    citecolor=black,
    filecolor=black,
    linkcolor=black,
    urlcolor=black
}

\definecolor{DarkerGreen}{RGB}{0,179,45}

\definecolor{Code}{rgb}{0,0,0}
\definecolor{Decorators}{rgb}{0.5,0.5,0.5}
\definecolor{Numbers}{rgb}{0.5,0,0}
\definecolor{MatchingBrackets}{rgb}{0.25,0.5,0.5}
\definecolor{Keywords}{rgb}{0,0,1}
\definecolor{self}{rgb}{0,0,0}
\definecolor{Strings}{rgb}{0,0.63,0}
\definecolor{Comments}{rgb}{0,0.63,1}
\definecolor{Backquotes}{rgb}{0,0,0}
\definecolor{Classname}{rgb}{0,0,0}
\definecolor{FunctionName}{rgb}{0,0,.7}
\definecolor{Operators}{rgb}{0,0,0}
\definecolor{Background}{rgb}{0.98,0.98,0.98}

\lstdefinestyle{python}{
  numbers=left,
  numberstyle=\footnotesize,
  numbersep=1em,
  xleftmargin=1em,
  framextopmargin=2em,
  framexbottommargin=2em,
  showspaces=false,
  showtabs=false,
  showstringspaces=false,
  frame=l,
  tabsize=4,
  basicstyle=\ttfamily\small\setstretch{1},
  backgroundcolor=\color{Background},
  language=Python,
  commentstyle=\color{Comments}\slshape,
  stringstyle=\color{Strings},
  morecomment=[s][\color{Strings}]{"""}{"""},
  morecomment=[s][\color{Strings}]{'''}{'''},
  morekeywords={import,from,class,def,for,while,if,is,in,elif,else,not,and,or,print,break,continue,return,True,False,None,access,as,,del,except,exec,finally,global,import,lambda,pass,print,raise,try,assert},
  keywordstyle={\color{Keywords}\bfseries},
  morekeywords={[2]@invariant},
  keywordstyle={[2]\color{Decorators}\slshape},
  emph={self},
  emphstyle={\color{self}\slshape},
  breaklines=true
}

\newtheorem{exmp}{Example}[section]

\newcommand{\splitatcommas}[1]{\begingroup\lccode`~=`, \lowercase{\endgroup
    \edef~{\mathchar\the\mathcode`, \penalty0 \noexpand\hspace{0pt plus 1em}}%
  }\mathcode`,="8000 #1%
  }

\tikzset{
>=stealth',
  punktchain/.style={
    rectangle, 
    rounded corners, 
    draw=black, very thick,
    text width=10em, 
    minimum height=3em, 
    text centered, 
    on chain},
  line/.style={draw, thick, <-},
  element/.style={
    tape,
    top color=white,
    bottom color=blue!50!black!60!,
    minimum width=8em,
    draw=blue!40!black!90, very thick,
    text width=10em, 
    minimum height=3.5em, 
    text centered, 
    on chain},
  every join/.style={->, thick,shorten >=1pt},
  decoration={brace},
  tuborg/.style={decorate},
  tubnode/.style={midway, right=2pt},
}

\makeglossaries

\newglossaryentry{non parametric}
{
    name=non parametric,
    description={Non parametric statistics does not assume that the underlying probability distribution has a predefined form, e.g., of type normal or exponential, and nevertheless is able to apply statistical inference to the problem. }
}

\newglossaryentry{control interval}
{
    name=control interval,
    description={For a random variable $X$, a control interval is a segment $[low, high]$ such that the probability of having a random variable materialising outside the segment is low}
}

\newglossaryentry{ML model}
{
    name=ML model,
    description={Depending on the learning task this may stand for different things.  In general a ML model some determining function that has predictive capability.   For example, in the case of a clarification ML task it will be a function that given a new objected, e.g., en image, 
determine its type, e.g., whether it is a cat or a dog. }
}

\newglossaryentry{bootstrapping}
{
    name=bootstrapping,
    description={Given a sample, $P$, from some distribution $F$, the empirical distribution of the sample $S$ represents the distribution, $P$, if the sample is big enough.  We can thus re-sample from $S$ with replacement to obtain fresh samples that approximate fresh samples from $P$.  This process is called bootstrapping.   }
}

\bibliography{main}

\begin{document}

\title{%
Theory and Practice of Quality Assurance for Machine Learning Systems \\
\large{An Experiment Driven Approach}
}

\author[1]{Samuel Ackerman}
\author[3]{Guy Barash}
\author[1]{Eitan Farchi}
\author[1]{Orna Raz}
\author[2]{Onn Shehory}
\affil[1]{IBM Haifa Research Lab}
\affil[2]{Bar Ilan University}
\affil[3]{Western Digital}
\date{2022}

\maketitle
\pagenumbering{gobble}
\newpage
\pagenumbering{roman}
\tableofcontents
\newpage
\pagenumbering{arabic}

\begin{abstract}

This book teaches the art of crafting and developing ML based systems.
Crafting machine learning  (ML) based systems that are business-grade and can be used by a business is highly challenging. It requires statistical control throughout the system development life cycle. 
This book advocates an "experiment first" approach stressing the need to define statistical experiments from the beginning of the system development life cycle.   
It does so through introducing methods for careful quantification of business requirements and identification of key factors that impact the business requirements. This reduces the risk of a project or ML system development failure.    The quantification of business requirements results in the definition of random variables representing the system key performance indicators that need to be analyzed through statistical experiments.  In addition, available data for training and experiments impacts the design of the system.  Once the system is developed, it is tested and continually monitored to ensure it meets its business requirements.  This is done through the continued application of statistical experiments to analyze and control the key performance indicators.   
This book also discusses in detail how to apply statistical control on the ML based system throughout its lifecycle. 
\end{abstract}

\chapter*{Thanks}
\label{Chapter:thanks}

We thank Guy Barash for working out solutions for the programming and experiments exercises, Krithika Prakash for commenting on the early text and discussion on the exercises, Diptikalyan Saha for discussion on the unit test vs. system test chapter, Radha Ratnaparkhi for discussion and insights on the unit and system testing concept, and Rakesh Ranjan for discussion, feedback and insights on the early versions of the ML pitfalls material.   
\chapter{Introduction}

The last decade has seen unprecedented growth in the deployment of Machine Learning (ML) based solutions, 
broadly influencing many aspects of our lives \cite{CNN2019}. 
Significant progress has been made in machine learning tasks such as image recognition, language translation, and tasks related to self-driving cars, to name a few. With such prospects of potential success, the industry is attempting to apply ML to a variety of use cases at an unprecedented rate, including crime prevention, medicine, IT and cloud analysis, cyberspace security, intelligent chatbots, and many more. 

ML-based systems combine ML models and regular software to achieve some common goal, e.g., making the driving decisions of the self-driving car. But what are ML models? ML models can be thought of as regular computer programs that were produced by automatically searching over some predefined set of possible programs.  The search is guided by examples of the desired program behavior.  We usually refer to the examples as the training data or simply as data.  Thus, just as we teach young children by example, in the case of ML models, we provide the machine with examples from which it learns the correct behavior; hence, the name machine learning.

ML models are statistical by design. We do not expect to find an ML model that is always correct.  In fact, we are unlikely to know whether the space of possible models that we search in contains such a perfect model, and even if it does it may be too difficult to find.  Instead, we settle for the model being correct most of the time.  In addition, we require that a successful model will be correct most of the time on unseen data of the same type as the data it was trained on.  
This trait is commonly referred to as {\em model generalization}. Thus, a bug in an ML model usually cannot be determined from the individual response of the model to a single data point; rather, it is determined by the aggregated behavior of the ML model on many examples. As a result, an ML solution composed of several ML models and regular software can only be successfully crafted if it is validated by statistical experiments.   It also means that the required behavior of an ML solution can only be defined as an aggregate of its behavior on individual data points. Therefore, the control of the aggregated statistical behaviour of the crafted ML system as a whole is indeed a major challenge addressed in this book. The average aggregate behavior and its control can be exemplified in the case of a self-driving car. For example, we can specify that the desired average number of wrongly recognized type of vehicle is no more than 5\%.

With this in mind, we emphasize the need for well-defined quantitative measures that reflect the desired behavior of the ML-based system. An example of such a measure is the average number of service calls a chatbot, whose goal is to address those calls autonomously, is routing to a human service agent. Such measures can be derived from {\em statistical functional requirements} hence denoted {\em statistical functional measures} or SF measures in short. Once SF measures are properly defined, one can utilize them to analyze the desired behavior of the system. As described in this book, this analysis can be performed via appropriately designed statistical experiments that estimate the expected average and variance of the SF measures. A good experiment should provide accurate estimates of the SF measures when the ML-based system is deployed. This practice of defining SF measures and designing experiments that test them throughout the life-cycle of the ML-based system tackles the challenge of crafting an ML-based system that meets its requirements. 

Collecting a set of training examples for a given SF requirement is challenging too. The data may be lacking factors without which the desired model cannot make correct decisions. We might have too few examples, or examples that are not representative of the type and distribution of data that the system will encounter when deployed. 
Finally, in order to guide the learning, the data needs to include the "correct answer" as well as indication that it is indeed correct. Such indication is typically provided 
by labeled data, in which case the problem is a supervised learning problem. However, it often happens that an organization holds a representative set of examples that is readily available but are not labeled.  

As suggested above, this book aims to address the challenges of crafting robust ML solutions that meet their design requirements and in particular SF requirements, and hence can be confidently deployed. We focus on an experimental approach to the crafting of ML solutions. This approach can be thought of as the ML equivalent of the test-first approach applied to regular software. To emphasize this similarity, we name our approach {\em experiment-first}. The focus of this approach is on crafting, validating, and continually monitoring ML-based solutions via a series of carefully designed experiments at different levels of abstraction. The book is self-contained and addresses the mathematical and statistical techniques needed to apply the experiment-first approach.  It contains numerous practical programming as well as theoretical exercises and their solutions. Exercises are an integral part of the text and are designed to facilitate understanding of the ideas, concepts and methods introduced. 

The main target audience of this book is data scientists who craft ML-based solutions, architects who define what the solutions need to do, and testers who need to validate that the system actually works and meets its specifications. The book also attempts to bridge communication gaps between the stakeholders involved in the development of ML-based solutions. The experiment-first approach introduced in this book facilitates an end-to-end, unambiguous, quantitative definition of the steps to be taken in the development of an ML-based system. Our approach details the activities performed, and the artifacts consumed and generated, throughout the development and deployment of the solution. 
The book introduces the statistical experiment and its results as the language by which the stakeholders communicate effectively to craft the ML-based solution.        

A preferred way to design and implement experiments entails following a scientific experimental methodology. This methodology includes the definition of a problem, the formulation of a quantitative conjecture, and the design and implementation of an experiment that validates or rejects the conjecture.  For example, a problem could consist of finding the speed at which a ball will touch the ground when dropped from a high tower.  The conjecture is that the speed is proportional to the time, and the experiment involves dropping balls from towers of different heights and measuring how long it takes them to hit the ground. 

This book is organized as follows. In Chapter \ref{Chapter:scientific-method} we paint in broad strokes the scientific experimental methodology and highlight its applicability 
to ML system development and validation. As mentioned in the previous paragraph, following the scientific methodology includes defining a problem, making a conjecture about how the problem may be solved, and then designing and implementing an experiment to validate the conjecture.  When crafting an ML-based solution, the standard conjecture entails generalization, namely, that the system will behave as required on new inputs not seen before.  The problem that the conjecture implicitly aims to solve is being able to construct an ML-based system that meet the requirements on new data.  After defining the requirements we address specifics of how to formulate a generalization conjecture and how to craft the appropriate experiment to validate it and analyze the results in the event that the conjecture was wrong. We also discuss 
the way in which the results of the experiments influence system design and contribute to the continual monitoring and validation of the system throughout its life-cycle.  

Then, Chapter \ref{mindDirect} details practical aspects of the experiment-first methodology. They are described as a set of best practices and pitfalls that aim to quickly reduce the risks associated with the development of the ML-based system and help address the most important concerns first.  Practices proposed in Chapter \ref{mindDirect} should allow stakeholders to better validate, from day one, the potential for successful generalization given the business requirements and data. The practices should additionally identify optimal usage of information extracted from the data and application of the most suitable ML techniques to the problem at hand. 

When analyzing ML models, the focus is sometimes on generic performance measures, e.g., the accuracy of the model. In contrast, an ML-based solution has unique, specific SF requirements. For example, a service chatbot may be required to autonomously handle at least $90\%$ of its conversations with customers, and send at most $10\%$ of to a human service agent for processing. In addition, specific requirements are frequently realized through the interaction among ML models and regular software.  Chapter \ref{unitSystem} discusses the composition of ML software and regular software in a unified ML-based system, and how such composition impacts system validation.

The next two chapters on ML testing before deployment (Chapter \ref{Chapter:MLtesting}) and on drift (Chapter \ref{Chapter:drift}) address the system correctness concern at different phases of the system life cycle. The ML testing chapter focuses on checking system correctness before it is deployed, while the chapter on drift detection analyzes whether the system operates correctly once it is deployed.  

Generalization is a fundamental conjecture made about ML-based system correctness. 
Thus, the experiments performed at testing time should indicate 
that the desired behavior will manifest once the system is deployed. Due to the stochastic nature of ML-based systems, the validation typically includes a stability assertion 
An assertion could suggest, e.g., that on average, 5$\%$ of the service calls received by a service chatbot are diverted to a human service agent, but it is unlikely that more than 10$\%$ of the calls are diverted. 
As data changes over time the system might fail to meet the desired behavior once it is in operation, even if it was well built and tested. For example, in the chatbot scenario, changes in the type of service calls might result in much more than 10$\%$ of the calls being diverted to a human agent. To handle this, drift detection is required. This topic is addressed in Chapter \ref{Chapter:drift}.  

Once the expected system behavior is established, we can define an optimal business process. 
In the chatbot example, once we know that up to 10$\%$ of the service calls are diverted to a human agent, we can plan the size of the customer service team that will provide proper response to the diverted calls. The integration of ML-based system behavior with the business process is discussed in Chapter \ref{optimal}. Finally, 
in Chapter \ref{industrial}, we present an industrial example to demonstrate the implementation of principles of the experiment-first methodology in practice.
Solutions to exercises that appear across the book. Code samples and mathematical background are found in the appendix. 

This concludes the description of the content of this book. It is important to note that the book does not cover topics such as generic trust-related measures and MLops. 
ML models learn from data. Hence, their inferences may be affected by data bias and adversarial manipulation. 
Some ML models can be trusted only if their decisions are comprehensible. 
Much research addresses these concerns \cite{trust}, \cite{RSG2016}.
%
%
Low-level engineering concerns (e.g., automatic model re-training) are out of scope too, 
and are well covered in the art (e.g., MLops tutorial \cite{MLops}). 

\chapter{The Scientific Approach to AI analysis}
\label{Chapter:scientific-method}
The greatest risk of a ML project is whether or not we can learn, i.e., build good ML models from data.  Typically, training data will also serve as input to the ML solution once it is deployed. The fundamental assumption of ML theory is that the data used for learning is representative of the data that will be encountered once the ML solution is deployed.  Thus,  the fundamental requirement from a good learning system is generalization - the ability to create rules that will work on data input of the same sort.  For example, a ML model that was trained to identify a human face on data that includes all types of animal and human faces is not expected to work well when presented with a picture of a car.  As typical in software engineering, the biggest risk of a project should be addressed first.   Thus, we should first address the question of whether, given the available data and the business requirement, we can successfully learn?   To address this concern, we design experiments whose results indicate how we can learn, if at all, the type of decomposition to components or architecture the system should have, and the next development iteration and experiments.  

\begin{example}
Our ML solution needs to determine the risk level an investor desires and different major expenses she is about to encounter, such as paying the kids' college tuition, buying a new house or going on annual vacations.  The available data is the investor's personal profile, past investment history, and assets inventory.   There are many possible architectures.  We can create one monolithic ML model.  In contrast, and on the other end of the spectrum, we can build a ML model that tries to predict the desired investment risk level and a model for each possible major investment, e.g., buying a house.  We can also have a single model for all of the expenses.  Finally, the models output that predict the expense types can be used as input or features in the learning of the risk level model.  Chapter \ref{mindDirecting} will provide insights into the tradeoff involved in choosing the right architecture, but the ultimate check is an experiment in which the different approaches are compared and the best choice is determined.  
\end{example}

\begin{exercise}
A self-driving car is trained using data on traffic in Tokyo.   The manufacturer claims that the car can drive in large European cities as they are similar to Tokyo.  Form an opinion about the claim and discuss relevant factors that you think impact the self-driving car's performance.  What  experiment might validate the manufacturer's claim?
\end{exercise}

It is useful to draw an analogy with the Test First software development process. In Test First, a functional test is written before the code is implemented.  The test formally defines what the system is required to do.  The code that is implementing the functionality captured by the test is then implemented.  Next, the test is executed against the code.  If the test passes, then the code is OK.  The test is added to a set of passing tests that accumulate and serve as a functional test suite of the system being developed.  At every development stage, all tests in the test suite must pass before the next test is designed.

In the case of a ML solution, an experiment is defined instead of a test. The experiment
results are used to determine whether the system meets the requirements, but with two
major differences: (1) the experiment result is a statistic, e.g., an average, and not a
success or pass binary result; and (2) the experiment result may help us choose among
design alternatives.

\begin{exercise}
\label{estimatePrecentage}
Given a number, x, a function returns $x^2$ most of the time, i.e., in p percentage of the cases.  Write a test that finds the percentage p.  How is that different from a test that checks that a deterministic function correctly evaluates $x^2$ ?

\href{https://colab.research.google.com/drive/1uO8CDjovgLwVng2rVFnfri9tPKWJBBAo?usp=sharing}{Click here for a Solution.} or see code in \textit{Solutions} appendix \ref{estimatePrecentage_sol}.
\end{exercise}

\begin{exercise}
Test First goes hand in hand with automation; we require that the test suite can be run and its pass or failed results determined programmatically.  What is the analogous requirement in the case of an experiment-based approach to ML development? 
\end{exercise}

We conclude that a good ML project consists of a series of statistical experiments.  The results of each experiment reduce the risk of the project and help determine its architecture.  It is similar to a sail boat that depends on the direction of the wind  -  the data in our case is the wind.  This approach, which we will call Experiment First, focuses on simultaneously considering the data \textit{and} the business goal from day one. We next take a deeper look at the design and analysis of statistical experiments and how they relate to the practice of Experiment First.

TODO:  add a general mapping of the scientific approach to AI analysis to setup the stage for the more detailed discussion in latter chapters in second part of the book.

\section{The scientific method}

The scientific method includes the consideration of a problem, the formulating of a conjecture, the design of an experiment that can validate or refute the conjecture, and then the analysis of the results of the experiments.   Insights from the analysis of the experiments lead to the formulation of new conjectures and new experiments, and the process repeats itself.

The scientific method is general.  It applies to problems such as finding a cure for a disease, discovering the laws that govern the movements of bodies, or determining whether a production line is reliable. The experiments used to validate the conjectures we have made with regard to such problems are designed using statistics as an underlying technique. We will explore the design and analysis of statistical experiments with a focus on ML, but first we will consider a simple example of the scientific method from mechanics.

\begin{example}
The problem is to determine the rule that governs "free fall" on Earth and the Moon.  We drop a small metal ball from a height of 2 meters and use a camera to determine where the ball is after 1 ms, 2 ms, etc. We repeat the same experiment on Earth and on the Moon.  Analyzing the results, we see that the location at time t, $x(t)$, is given by a formula of the form  $x(t) = At^2$, where A is a constant that has different values on the Moon and on Earth.   We are happy and declare victory - in determining the rule of "free fall".
\end{example}

\begin{exercise}
Can you suggest why the above conclusion is not correct.  Clue - think about letting other objects fall.   
\end{exercise}

It turns out that designing good experiments is tricky in that it requires domain knowledge: we need to think about the factors that influence the results of the experiment, which is hard to do if we lack that knowledge.  In the experiment above, factors such as wind or the nature of the object being dropped, e.g., a feather instead of a metal ball, are applicable to the Earth but not to the Moon.  Similarly, domain knowledge is key to the successful design of ML solutions.  This is common knowledge, often expressed through feature engineering, among other things. Feature engineering is the practice of designing the features extract from the raw data that will be used in the learning process. What is less obvious is that domain knowledge is also the key to the analysis and validation of ML solutions.  The Experiment First approach emphasizes this point. In this approach, domain knowledge is embedded in the validation process through careful design of the experiments and identification of the factors that govern them.  We next consider the relation between the scientific method and the ML solution development process.

\section{Mapping the scientific method to ML development}

As discussed in the previous section, following the scientific method includes defining a problem, coming up with a conjecture and the design, implementation and analysis of an experiment that confirms or rejects the conjecture.  In the case of ML embedded systems the problem is expressed in the form of a business requirement.  For example, the bank would like to develop a system that determines if to give a loan to a customer or not.  The conjecture may be that the personal and past data on the customer financial transactions is enough to create a ML that determines if the bank can make a profit from giving the loan.  The experiment will constitute determining the performance of the ML on a new and representative set of customers.  In other words, checking if the ML model was able to determine correctly from the personal and financial transaction data of a fresh set of bank customers that the model has never seen before if the bank will profit from giving the loan or not.  

Key points in the design and analysis of the experiment include translating the business objective to quantitative measures that can be measured in the experiment, determining the factors that are impacting the experiment, insuring coverage of these factors, and statistically analyze the experiment outcomes to reliably deduce results from the outcomes.   Error and randomness should be taken care of when analyzing and interpreting the results. The last point is illustrated by the next thought exercises.

\begin{exercise}
We consider the tossing of two coins 100 times.  This is done for two coins.  The first coin turns head 80 times and the second coin turns head 85 times.   Is the second coin more likely to turn head than the first one?  Discuss why it may be the case that it is not.  How would you go about increasing your confidence that the coins have a different chance of turning head? 
\end{exercise}

\begin{exercise}
The bank also has data  on whether or not its customers paid previous loans.   Using that data customers who previously paid on time are labeled as  customers that  "can be given a loan" but if they did not pay on time they are labeled as "can not be given a loan" .    Only using data available before the loan was given, two  ML models that predict if the customer can be given a loan are developed.  Next the percentage of correct  answers the models gave  is calculated on a fresh set of customers that were given a loan.   One model predicts correctly 80 percent of the time and the other 85 percent of the time.   Which model is better and how is the answer to the question related to the previous question with the coin tossing?
\end{exercise}

It is important to note when analyzing ML embedded systems that randomness is ingrained in the  solution.  If we use the two models developed for loan prediction in different time windows we expect percentage of the correct  answers to change. We are thus interested in whether or not most of time it is likely that the percentage of correct  prediction of one model is going to be larger than the  percentage of the correct prediction of the other model.   This is a statistical question and requires statistical techniques we will develop in chapter \ref{Chapter:MLtesting}.

Two other points are interesting to note when considering the bank loan example.  The translation of the business requirement to a quantitative measure is a non trivial part of the system definition as well as the experiment definition.  In our case, a customer could be late, but not too much, and still the bank may make a profit on the loan.  Thus, other more refined labels could be defined.  In addition, there are customers that were never given a loan.  Maybe such customers are of different "type"?  Maybe our prediction will not work well on customers that were never given a loan?  Consider the following thought exercise. 

\begin{exercise}
Can you suggest a new experiment that will help distinguish if the models have different performance on customers that were never given loans before?  What would be a conjecture in this case?
\end{exercise}

We could distinguish between customers that were given a loan only once and customers that were given a loan more than once. We can then take the label for loan payments for each loan the bank gave to the customer separately.  We can thus have a model for customers that were given the loan for the first time and design a second experiment that determines if the model work well on  such customers.  In fact we have identified a factor that governs the experiments, namely, whether or not customers are given a loan for the first time and an implicit requirement - the model should preform well on customers that are given a loan for the first time as well as on customer that were previously given a loan.  Such a factor identification leads to an architecture question - should we have two models one for customers that were never given a loan before and one for customer that were?  

\begin{exercise}
How would you determine which architecture option to take?
\end{exercise}

We have seen how business requirements are translated into quantitative ML embedded systems performance requirements on fresh data.  We also seen how an experiment designed to validate initial models lead to the identification of factors governing the experiment, the identification of not yet articulated business requirements and the design of additional tests to validate the evolving ML embedded solution performance and determine its architecture.  Next, we will discuss each stage of the scientific method in the context of ML in more details.  

A comment on randomness is in order.  When we say that we choose a fresh sample, we mean that we randomly choose a new set of customers.  If the choice is skewed, for example with more customers that were previously given a loan, then the percentage we obtain may be skewed and not reflect the real behavior of the system.  The important point is that randomness will take care of factors that impact performance even if we have not yet identified them -  a fundamental principle of statistical experiment design that applies to the crafting and analysis of ML solutions.  We will revisit this point in various ways throughout this book. 

\section{Conjectures}

We next continue in the discussion of the relation between the scientific method and the development of a ML embedded system and recall that the scientific method includes the following steps - problem definition, conjecture formulation and experiment design.  The problem is defined by the system requirements and the conjecture is driven from them.  Deriving the conjecture includes the translation of the system requirements to a quantitative measure and the identification of factors that govern the quantitative measure and desired performance expressed using this measure.   Typically, the ML conjecture is that independent of the governing factors the quantitative measure desired performance can be learned. Learned here means that the ML embedded system 
will preform according to the desired level as measured by the performance measure on new data that was never used in the training of the ML models embedded the system.

Interestingly the step of quantifying the system requirements and the factors that govern them help identify system requirements that are not a good candidate to be implemented as a ML embedded systems.   Generally speaking systems could be too complex or too simple to be good candidates for a ML emended implementation.  If a close form of the relation between the governing factors and the desired performance can be expressed and only the optimal parameters that define the relation obtained, the desired system is a too simple to be a good candidate for a ML embedded implementation.  The example of the law of mechanics that were given above is a good example of such a system.   On the other hand, if the governing factors are hard to identify and it is not clear of what type the relation that implements the system is the system is too complex and will probably not be a good candidate for a ML embedded system implementation.  An example of that is a long term accurate weather predication systems.  There is a sweet spot in between of the two extreme which serves as a good candidate for a ML embedded solution which we discuss next.

A good ML solution candidate will at least meet the following criterion. 
\begin{enumerate}
    \item We are able to quantify the business requirements.
    \item The factors that govern the business requirements can be identified.
    \item We identify a search space in which we can search for a good solution automatically.  This is sometimes referred to as the learning phase and the search space is referred to as a model (e.g., a regression, a neural network with some specific architecture or a decision tree).
\end{enumerate}

The last requirement is many times met implicitly by choosing a ML learning method that previously worked well in some domain, e.g, convolution neural network (CNN) for anomaly detection. Making it an explicit step increases the chance of successful learning.  In addition, autoML \cite{???} can help automate this step.  The following exercise is designed to help clarify the concepts of a model and model choice. It requires experience in the use of ML libraries and training.  If such experience is not available it is recommended that the reader will go through some tutorial on linear regression using Python before attempting the exercise.  

\begin{exercise}
\label{Chapter:MLtesting:Excercise:parabola}
A parabola, i.e., $f(x) = ax^2+bx+c$, is given. 100 samples $D = \{(x_1, f(x_1)), \ldots (x_n, f(x_n))\}$ are obtained by randomly choosing $x_1 \ldots x_n$ and then calculating $f(x_1), \ldots, f(x_n)$.  D is used to train a linear regression model of the form $g(x) = dx+e$. 
\begin{enumerate}
    \item Do you expect the results to be good?  Explain your reasoning.  
    \item Implement the above scenario using standard learning libraries.  Validate whether or not the obtained linear regression was good.
    \item Provide a graph view that presents $f()$ and the linear model obtained by applying the regression on D.  How does that help understand the performance of the linear model?  
    \item Is there any pair of parameters d and e of $g() = dx+e$ that will work well?
    \item Use the following feature engineering.  Define $D^{'} = \{ ((x_1,x_1^2), f(x_1)) \ldots ((x_n, x_n^2), f(x_n))$.  Apply linear regression on the new data set.  Did the results improve?  Can you explain why?
    \item What will happen if you add some noise to the data?   Concretely, randomly choose some $noise_i$ from the standard normal distribution, $N(0, 1)$, $n$ times and change the data to $D = \{(x_1, f(x_1) + noise_1), \ldots (x_n, f(x_n) + noise_n)\}$.  How would your answers change given the new data set?
    
    \item Does the size of the dataset $D$ matters? if so, how?
\end{enumerate}

\href{https://colab.research.google.com/drive/15Quc4KJ9CTVBIui3-ylCQpcXo8lR0DgY?usp=sharing}{Click here for Colab Solution} or in the solution appendix \ref{Chapter:MLtesting:Excercise:parabola_sol}

\end{exercise}

TBC - explain how the parabola example is related to the model concept.

TBC 

\section{Experimental design}

We have discussed derivation of a quantitative performance measures from business requirements and the identification of factors that impact the performance measure.   In the language of statistical experiments a performance measure is referred to as the dependent variable and the factors that impacts the performance measure are referred to as the independent variables. Our implicit conjecture is that regardless of the value of the dependent variables the performance of the ML embedded system on new unseen data will meet the require performance level as defined on the performance measure.   For example, the performance level may require that the average percentage of wrongly predicting that the bank will make a profit when giving a loan to a customer is 10\%.  In addition, its variance is required to be small.   

In order to conduct the experiment, we will 
typically collect a set of input data and apply the ML based system on the data to obtain an estimate on the performance of the dependent variable. We would like the identified independent variables to appear in the data. In reality one can not expect to identify all of the independent variables.  We thus expect to have some latent variables, possibly with their value appearing in the data, that impact the dependent variable we are estimating.  Unfortunately, the dependent variables may not even be in the data.  
  
Denote the dependent variable by Y, the identified independent variables by $X_i$ and the unidentified independent variables by $X^l_j$.  As we collect our data for our experiment we need to take into account the following.

\begin{itemize}
\item Make the data records as comprehensive as possible.  Include available features in the data record, e.g., age of the customer, if available even if it is not required as input to the ML based system and it is not identified as an independent variable.  This will allow for further analysis aimed at discovering the latent variables $X^l_j$.
\item Avoid bias.  Avoid collecting the data in a way that consistently favour some data record features.  One of the ways to mitigate the issue of latent variables is to randomly sample the experiment data.   If this is not done fundamental assumptions underlying the theory and practice of ML are broken.  In addition, by randomly sampling the data the impact of latent variables $X^l_j$ appearing in the data is amortised and we can still estimate the behaviour of the dependent variable Y although we did not identify the latent variable.
\item Choose whether or not to stratify values of dependent variables.  For example, we are given a medical system that decides if a new drug's side effect should be reported to the government based on medical reports.  An independent variable is the type of the report.  There are two types of medical reports - hand written reports and digital reports.  We may be interested in the system performance when the report type, X is $X = handWriten$ and when $X = digital$.  We will thus have two different predictions, one for each value of the independent variable X.  This is called stratification. On the other hand, we may choose not to stratify the independent variable.  That brings us back to the randomization guideline.  We need to take extra care that the variable is not biased.  For example, if our sampled data mostly have digital reports we can not expect to draw conclusions about the performance of the ML based system on hand written reports.   When do we stratify?  When we want the analysis and predication at that level of details to clarify if the system really meets the business requirement. the business requirement in our case is that the the system will work well on both values of the variable X (hand written and digital).   Note the obvious dilemma.  If we stratify all possible values of the independent variable $X_i$ we will get a combinatorial explosion of stratified values.  we will revisit this issue later in the chapter.   
\end{itemize}

The following exercise is designed to demonstrate the concepts of randomness and stratification.  

\begin{exercise}
The unknown relation the system had to learn is $f(X, Y, Z) = X + Y^2 - Z^3$.  X and Y take values in $\{-1, 1\}$ while Z is a number, $Z \in R$.  We identified the independent variable X.  Data was sampled as follows.  X was set to -1, Y was randomly chosen. Z was randomly chosen from a normal distribution with average 5 and variance 1. 
\begin{enumerate}
\item How is the data biased?
\item Suggest a correction to the sampling method
\item How would you conduct a stratification on X?
\item What are the hidden variables?
\item Assume that the system was implemented as $g(X, Y, Z) = Z + Y^2 - Z^2$.  Implement an experiment that analyze the performance of this system and suggest ways to identify that Z is a dependent variable.
\end{enumerate}
\end{exercise}

The following exercise is designed to shade light on sampling challenges used to collect the data.  

\begin{exercise}
The ML systems attempts to determine if a child will graduate from school.  Two independent variables are considered, namely, the child's neighbourhood and the size of the child's family.  It is desired that the ML system will have the same level of prediction regardless of the values of the independent variables and we are collecting data to measure and determine if this is indeed the case (the experiment).  Three sampling methods are considered as follows.
\begin{enumerate}
    \item Randomly choose a child
    \item Randomly choose the family size, then randomly choose a family of the chosen size, then randomly choose a child in the family that is attending school
    \item Randomly choose a neighbourhood, then randomly choose a family that lived in the neighbourhood in the last 5 years and then randomly choose a child in that family    
\end{enumerate}
Can you identify bias introduced by the sampling techniques above?  Assume the first technique produced a sample of 100 children and non comes from a family of more than 6 children. Can you still estimate the performance of the ML system?   What would you suspect if no child from a family of one child was chosen in the sample?
\end{exercise}

\chapter{Pitfalls and Best Practices for the Machine Learning Designer and Tester}
\label{mindDirect}

\section{Overview}

We are in the middle of the AI and data revolution.  Companies are attempting to utilize data and realize business value. 
"Data insight" is an old-fashioned application of AI that does not recommend actions.  However, businesses today want to trust AI in their decision making with and even without a human in the loop. This requires the creation of a business-grade reliable and trustable ML solution (or AI-infused solutions).  

There are many good resources on ML technology as well as good courses on theory and algorithms.  We are not attempting in this paper to fill that need; if we were, we would be presenting Yet Another Machine Learning Machine or YAMLM.   Instead we are trying to highlight the best mindset that increases the chance that an ML project will succeed.

\section{From day one determine if the data is relevant for the business objective}

Deciding which business objective to implement  depends on the data.  The happy marriage of business requirements with relevant data is the determining factor in creating a high quality AI-infused system. It's a waste of time to discuss the solution in isolation from the data and without directly inspecting and analyzing it to determine whether the business goal may be achieved.  However, it may be useless to learn whatever is possible to learn from the data as such learning, even if successful, may not serve any useful business goal.

Following items should be added/integrated TBC:

\begin{itemize}
\item
Data may or may not include significant signal supporting the buisness objective. For example, classification of software defects ODC (Orthogonal Defect Classification) vs. custom classification
\item
Data may have a signal that is ‘too strong’
Ex. It is always the case that a defect opened by a manager is of highest importance. No sense in learning this. Use a rule instead
\item
Explicitly measure business value; Don’t confuse ML metrics with business value.
\item
Explicitly measure business value - Don’t confuse ML metrics with business value 
\item
Customize off-the-shelf ML measurements for experiments

\end{itemize}

\section{Think science and experiments}

AI-infused solutions are heavily dependent on data and are statistically correct by nature: they are by definition sometimes correct and sometimes incorrect. It's the nature of the beast. Business-grade AI-infused solutions need to control the statistical error.  Experimental design approaches should thus be introduced and analyzed. These experiments are equated with ML learning iterations.   This will be further explained here.  To reduce the main risk of AI-infused systems development, frequent experimentation is desirable and architecture should be driven by the experimental results.  Consider a sailing ship attempting to reach point B from point A.   The wind is not going from B to A and needs to be harnessed for the boat to reach its destination. In our case, the wind is the data and the sailing ship is the AI-infused system under development and the harnessing process is realized through experiments.   

TBC:
\begin{itemize}
    \item Explain the scientific method
    \item Explain design of statistical experiments
    \item Map statistical experiments to machine learning
\end{itemize}

\section{Data must be of valuable volume; When not to use learning}

Systems scale in complexity from the ones for which we cannot even list the factors that control them to the ones for which we can easily write their equations, e.g., a physical body that accelerates under a constant force. In such a case use rules; ML may be overkill. An intermediate situation is where the equation is known but a few constants should be empirically calculated, e.g.,  system performance tuning.  Maybe a few cycles of experiments will do in such a case.  Yet another intermediary case where ML shines is when only a general form is known (a model).  We need to think of the choice of a ML technique as a choice of such a model.  For example, a neural network is a model with a graph of thousands of neurons, where the connecting edges are associated with weights.  Together the set of all weights is our model, and we are tasked with finding the best set of weights.

Even if we decided to apply learning, sufficient data is required to succeed. A rule of thumb for data sufficiency is that the size of the training data should be an order of magnitude bigger, e.g., at least 10x, than the number of features.  If a classification task is attempted, the number of examples per label should be at least 30. 

TBC:

\begin{itemize}
    \item is the training data representative of the deployment data?
    \item Remember that you also need data for testing
When training – development set – used multiple times 
When testing – test set – you cannot use it to change your model! If you do, you will need more data for testing
    \item Remember labels. You cannot learn labels that are not in the training set. Need to understand and control labeling mistake rate (see Estimate the probability of a labeling mistake)
    \item Is data adequate for learning? A hybrid approach
\end{itemize}

\section{Use all relevant data and information: Do NOT ignore structure}
\label{mindDirecting}

When designing an AI-infused solution, you will typically consider some desired business value and a given source of data.  For example, the business objective states that "given 1000 tickets (problem reports), try to determine the nature of the problem automatically".  Don’t accept the given data as the only source of information. Instead, explore other possible sources through the following heuristics.  

\begin{itemize}
    \item Exploiting implicit data
    \item Taking advantage of meta data associated with the data you were given
    \item Consider implicit or explicit tracability and represent the data in such a way that takes advantage of these relations
    \item Structure contains information.  Resist the temptation to lose structure in order to apply some ML algorithm. Maybe the data was exported and structure lost.   Try in such cases to reverse engineer the structure.  
\end{itemize}

We'll elaborate on these heuristics here.

TBC:

\begin{itemize}
    \item Explain the heuristics : "implicit", "meta data", "traceability" and "structure". 
    \item Use all relevant data: Do NOT ignore structured data!
\end{itemize}

\section{Is the training data representative of deployment data?}

TBC:

Assumption underlying statistical machine learning: the training and test data distributions represent the deployment distribution.  In fact, many times the data is collected in a totally different way than the data the system will encounter during deployment.  For example, if self-driving car is trained in different weather conditions or in different countries.  Another example is that data collected is different due to problems in measurements or different scales

How to handle:

Missing data - The scale is different (camera is positioned higher), Use knowledge about the data distribution to determine if the data meets reality

Domain knowledge:

If you know that percentage of fatal side effects is negligible and you get data that has 0.80 fatal side effects, then something is wrong with the data.  Another example of domain knowledge.  Breast cancer percentage in the population is around 0.5

Statistical relations on data:

Capture and monitor feature relations and behavior (min/max, correlation,…)
Hands-on:

Otherwise you need to design experiments so that you demonstrate that your model monotonically improves with new data as you know that the data will change over time and your data is not representative.   

\section{Get to a supervised learning problem}

ML shines when supervised learning can be applied.  It is more mature and has a stronger success record.  It’s also an indirect indication that you are able to quantify a business objective that will be used in testing the system.  Accurate labels are required for training and testing.  The challenge is to identify appropriate labels for the data.  

Two general heuristics apply:

\begin{itemize}
\item
Choose a subset of the attributes in the data for the labels.
\item
Create an experiment that will generate desired labeled data.
\end{itemize}

TBC:
To predict system performance, a labeled test set will be required. 

\section{Minimize manual labeling, instead utilize experiments, rules and boosting}

TBC - minimize remove hyper parameters.

\chapter{Unit Test vs. System Test of ML Based Systems}
\label{unitSystem}

Why do we need more than one model to build a system?   Possibly, we can always feedback the entire data to the machine learning algorithm and produce the desired system?  Let's consider the following example. 

\begin{example}
\label{whyDecompose}
Consider a system that identifies the type of vehicle in an image.  We desire a rough classification into land vehicles that requires a track and those that do not, boats, submarines and flying vehicles.   The images have reliable meta data that clearly state if the vehicle is a land vehicle or not.  We could create two possible systems. We could try and build a monolithic model by training a model using the images and their meta data.  Another option is to create a model, $M_1$, that identifies boats, submarines and flying vehicles assuming that the input images are not land vehicles and another model, $M_2$, that identifies if the images are land vehicles that requires a track or not assuming that the vehicles are land vehicles.  We will then compose the two models to get our system in the following way. We will check the meta data of the image to determine if the vehicle is a land vehicle or not.  We will then apply $M_1$ if the image is not a land vehicle and $M_2$ if it is.  Which of the options is better?  As we assumed that the rule is reliable and it seems reasonable to assume that it is stable over time we can probably assume that the second system is much better.  In addition, the learning task in the second case are easier as we have less labels to classify so we may achieve better generalization overall.   
\end{example}

\begin{exercise}
What happens in the two cases if the vehicle is amphibious?
Try different possible assumption on the meta data categorization and analyze how the two possible systems are likely to behave under your assumptions.  
\end{exercise}

The example above demonstrates why a hybrid system composed of ML models and deterministic rules may be the ideal architectural choice.  Another interesting aspect of system decomposition for ML based systems is that its design should be driven by the data available for learning.  To see why, consider a system we would like to develop that assists in diagnosing a patient. There are many medical conditions and most of them are rare. Thus, we may have just a few examples for most of the medical conditions.  As a result, it may prove impossible to train a ML model that successfully predicts that the medical condition is a rare medical condition.  There are simply not enough examples for the ML to generalize and predict in the case of rare conditions!  This is a special case of stratification or slices on the independent variables that do not have enough training examples.  For example, we may know that cities and neighborhoods are both independent variables, that will affect whether or not a child will graduate from school, but we do not have examples for some of the cities or neighborhoods for which we would like to apply the model. Thus, we probably cannot train a ML model to predicts on those combinations of neighborhoods and cities for which we do not have training examples. Instead, the general rules is that slices for which there is not enough training data should be handled in the old fashion by developing deterministic rules that apply to them.  Again, the result is a hybrid system.

\begin{exercise}
\label{excercise:402}
We know that for some $a \in R$ the system is of the form $f(x) = Ax^2$ for $x > a$ and $f(x) = Bx$ when $x \leq a$.  You are given a training set $D = \{ (3, 9), (4, 16), (5, 25), (6, 36), (7, 49)\}$ where for a given point $(x, y)$ in the data set we know that $f(x) = y$.  Which of the following parameters $a, A, B$ can be learned and why?.  Next, assume that $a = 0$, $A = 1$ and $B = -1$. Attempt the following:
\begin{enumerate}
    \item Write a program that randomly generates data for the above case.  
    \item Implement a learning algorithm that will learn $a, A, B$ in this case.  Generalize your data generating process so that it will randomly generate legal data for random choices of $a, A, B$.  Show that your algorithms learns for a random sample of data generated by your data generating program.  
    \item Is your ML algorithm a composition of several ML algorithms or a monolithic solution?  Why did you chose one solution over the other and what were the trade offs that you took into account? 
\end{enumerate}

\href{https://colab.research.google.com/drive/1TlQUi7BaUvwXx3MOvyuIhLKmuc98U3x5?usp=sharing}{Click here for solution} or see solution \ref{excercise:402_sol} in  appendix \ref{chapter:solution_appendix}.
\end{exercise}

For regular systems components or units provide a well defined deterministic interface and they interact at the various levels of the system, create composed components and eventually obtain the desired system objective. Thus, testing for regular software consists of validating the expected deterministic behaviour of the system at each level of components composition. 

For ML based systems, testing consists, in addition to traditional testing, of experiments on random performance variables that needs to be controlled at the different system levels. In addition, the random performance variables need not be the same at all levels of the system. We may thus be interested in the accuracy of intent classifiers of the chatbot solution at the unit level but the average number of service calls that are directed to a human agent at the system level (see chapter \ref{industrial} for details).  Thus, experiments need to be conducted at the unit and system levels to establish control over their their associated and typically different random performance variables. In general any component of the system that has at least one ML model embedded in it will require validation of control of some random performance variables through appropriate experiments.  

As mentioned above, an interesting curiosity of the process of crafting ML based systems is that the optimal architecture is "driven by the data".   Thus, experiments not only serve to test the system but also to design it. In addition, the line between testing and designing is blurred.  We have seen how weakness in training data for slices of the data may drive the decomposition of the system.  Such weaknesses are discovered through experiments. Thus, indeed, the results of the experiments and the data available drive the design of the system. 

It is interesting to note that some view test first for regular software as a design paradigm. From that perspective the analogy between test first for regular software and experiment first for ML embedded systems is strengthen as both approaches are said to drive the design of the system.  

\begin{exercise}
Consider a complete rooted binary tree of depth n (i.e., a path from the root to a leaf has n vertices).  Each node in the tree represents a decision the software is making.  In addition, assume the decision at node $v$ of the tree has error probability $p_v$.   An adversary chooses a path, $v_1, \ldots, v_n$ from the root $v_1$ to a leaf $v_n$. A decision is then made by the system by making a decision at each node of $v_i$ with error probability $p_{v_i}$.  The system makes a correct decision only if all decisions along the path are correct. 
\begin{enumerate}
    \item What is the probability of making a correct decision if $v_1, \ldots, v_n$ is chosen by the adversary?
    \item Which path should the adversary choose?
    \item Implement an algorithm that finds the path the adversary should choose.  What is its running time and is that the best running time possible?
\end{enumerate}
\end{exercise}

A useful abstraction that applies to many ML based systems and will help us better understand the fundamental challenges of crafting ML based systems is a decision tree.   The system is using the decision tree to make decisions.  The nodes of the decision tree are decision point and can utilize either some deterministic rule or a ML model.  If the ML model is used at a decision tree node to make a decision, the decision is correct in probability and its correctness probability may also dependent on the independent variables that govern the system.   We refer to such a decision tree as the ML system hybrid decision tree.  For a detail example of such a system see the chatbot example in \ref{industrial}.  In that example the system uses intent classifiers to decide what the customer is interested in (paying a bill, withdrawing money from their account, etc), the system then proceed to follow a deterministic decision process that obtain the necessary data from the customer and complete the required service.  Given a system with a hybrid decision tree, we can now elaborate on the challenges encountered in testing it as follows.  We will refer below to nodes that use ML models to make a decision as non deterministic nodes and node that use rules to make the decisions as deterministic nodes.

\begin{enumerate}
    \item Each non deterministic node performance dependent variables needs to be identified and statistically controlled through an appropriated experiment. Note that there could be more than one dependent variable to analyze per non deterministic node.  For example, if the decision is whether or not to report some new drug side effect to the government, it may be more important that we report a true new side effect than that we do not report the new side effect when there is one. Thus, the two possible error types, namely reporting on a new side effect when it is not a new side effect or not reporting on a new side effect when it is a new side effect needs to be analyzed.\footnote{TBC - explain why you are not using standard terminology - false negative, false positive, precision and recall.  Not everything is binary..} 
    \item Is the hybrid decision tree the best way to design the system?   Can we suggest a different hybrid decision tree that will better utilize the data and get a better overall performance or a more stable one?
    \item Is the performance of the entire hybrid decision tree satisfactory?   What are the performance variables that need to be analyzed for the entire system?  These overall system performance variables need to be analyzed for each path in the hybrid decision tree that contain a non deterministic node. Some times we can deduce the analysis from the experiments on non deterministic nodes along a path (see proceeding exercises), but most likely we will need to conduct new experiments to test the entire path's performance.     
\end{enumerate}

The following exercises are designed to  clarify the concepts of unit test, system test, and system design using the hybrid decision tree for ML based systems. 

\begin{exercise}
Developers developed a classification model as part of a larger system and wanted to test it.  In order to do that they have obtained a new labeled data set, T, that is representative of the data that will be encountered when the model is deployed as part of the system and that was not used to develop the model.  They calculated the accuracy of the model, i.e., the percentage of correct answers the model gave on T, and the number was $95\%$ which seems to be a good number so they concluded that the model is validated.
\begin{enumerate}
    \item The probability that the model is less than $93\%$ accurate is required to be less than $0.05$.  Can we conclude that the model met this requirement?
    \item Is there anything wrong with what the developers have done?   Discuss your answer in light of statistical stability and system business requirements.  
    \item The classification problem is a binary calcification problem with two possible labels, -1 and 1.  T has 1000 data points 500 of label 1 and 500 of label -1.   On label 1 there are 10 mistakes and on label -1 there are 40 mistakes made by the model. What is the probability of mistake given that the label is -1 or given that the label is 1?  
    \item Every time the model makes a mistake on label 1 the company losses $100\$$ and when a mistake is done on label -1 the company losses $50\$$.   What is the expected loss from a 1000 data points encountered by the system? 
    \item The data in production drifted and now we expect 900 out of a 1000 records to be of label -1.  The model conditional probability of making a mistake on each of the labels (-1 and 1) remains the same.  What is the expected loss of the company on 1000 data point encountered by the system now?
\end{enumerate}
\end{exercise}

\begin{exercise}
To follow the experiment first approach, we use the following steps:
\begin{enumerate}
    \item Identify a requirement and quantify it resulting in a dependent variable we want to analyze through an appropriate experiment.  
    \item Use domain knowledge to identify the independent variables that impact the above dependent variable.
    \item Design an experiment that will predict the relation between the dependent and independent variables.  The dependent variable will be estimated using the result of applying the ML based system or any of its components on the system inputs.  Which part of the system is applied dependents on the level in which the test is conducted.  In the hybrid decision tree model the test may apply to a non deterministic node in the tree or to the entire tree.
\end{enumerate}

Consider the following questions.
\begin{enumerate}
\item Is there a difference in the statistical techniques that are applied to a non deterministic node in the decision tree or to a path in the hybrid decision tree?
\item Does it make sense to apply statistical tests to other parts of the hybrid decision tree (other than a path or a node)?
\item Consider your answer to the previous question.  How would you define unit test and system test in light of your answer?
\end{enumerate}

\end{exercise}

\begin{exercise}
Isaac Asimov defines his three laws of robotic in his 1942 short story "Runaround".  The laws read 
\begin{enumerate}
    \item First Law. A robot may not injure a human being or, through inaction, allow a human being to come to harm.
    \item Second Law. A robot must obey the orders given it by human beings except where such orders would conflict with the First Law.
    \item Third Law. A robot must protect its own existence as long as such protection does not conflict with the First or Second Law.
\end{enumerate}
Define a non deterministic decision tree that implements the above laws and is as "doable" as possible with the current state of the art of ML.
\end{exercise}

\chapter{ML Testing}
\label{Chapter:MLtesting}

As discussed in previous chapters ML based systems are becoming embedded in our everyday life. This chapter outlines how to confidentially engineer ML infused systems.  Indeed, it is challenging to build ML infused systems that we can rely on. Cutting to the heart of the matter: ML infused systems contain ML models. Such models are non deterministic and may sometimes make the wrong decision.  To handle the system inherit non determinism, we introduce statistical experimental design and explain how it is utilized to control the errors introduced by a ML infused system and thus make it reliable.     

This chapter is organized as follows.  We first introduce the problem of controlling ML infused system quality through an ideal and simple example and then deep dive into the design of statistical experiments and how they are used to develop, analyze and validate high quality ML based systems. At the heart of the chapter lies the identification of a random variable that represents a desired behaviour of the system and the analysis of its average and standard deviation.  This, in turn, let us define an envelope of control under which the system behaviour is acceptable.  The chapter builds on chapter \ref{Chapter:scientific-method} that discusses the design of statistical experiments to drive the development of ML based systems.

\section{Business grade ML solutions}

We are going to discuss the creation of business-grade ML solutions. The objective is to create an ML solution decision makers can use to guide their decisions with confidence or even let the ML solution make decisions without a human in the loop. This is not new.  For years statistics has been used to guide decisions and make them with confidence. For example, a production line is only profitable if up to $5\%$ of its produced items are defective.  We sample the production line every $20$ minutes and determine the percentage of defective items in the sample.  Apply a statistical test or a confidence interval that was previously developed we determine if production should be stopped and the production line re-tuned as there are too many defects making the current production not profitable.  The decision is made at a certain confidence level as there are probabilities associated with the two types of errors that we can make; The first is that the production line is profitable when it is not and the second is that the production line is not profitable when it is.


We make the following crucial observation.   ML solutions are a complicated version of the above production line example.  They are heavily dependent on data and are random by nature.  In other words, if a fresh training data sample is chosen the resulting ML model and its performance will change.  Thus, repeating the learning process on two different samples of training data will probably produce two slightly different ML models.  Also by definition sometimes ML models are correct and sometimes they are not correct! It's the nature of the beast.     

In order to achieve business-grade ML solutions we need to statistically control the solution error.  The following is meant by statistical control.  The solution performance, say accuracy, is a random variable.  We would like to be able to claim the following type of claim: with probability of error of no more than $5\%$, the accuracy of the solution lies between $84\%$ to $87\%$.
We call this interval a control interval as this interval is used to "control" the ML performance and determine if it is within the expected bounds.  

In other words, we want to be able to say what the expected performance of the system will be and quantitatively determine the probability we will make a mistake.  If we are able to do that we will also be able to determine if a ML solution is performing correctly in the field as we should rarely, if ever, see a performance that is outside the control interval.

Our approach as mentioned before is \gls{non parametric} as for modern ML infused systems it is hard to determine a distribution family and fit a distribution to the data.  Another motivation to the use of \gls{non parametric} approaches is the availability of modern computers. Utilizing the computation power of modern computers we will be able to rely on approaches such as \gls{bootstrapping} and Monte Carlo\footnote{Eitan - explain what Monte Carlo is.} to develop our ML infused system's control.  

We will start by studying how to estimate any random variable distribution in a \gls{non parametric} way.  This will serve as the foundations for our first ideal control interval example.  

For those not familiar with parametric confidence intervals it is recommended to review \ref{parametricControl} at this point.

\section{Random variables and empirical distribution}
\label{randomVariable}

We are given a probability space $(E, \Omega, P)$, $\Omega \subseteq P(E)$.  E is the set of possible events that can occur when we run our experiment.  Sometimes E is referred to as the set of elementary events to distinguish from subset of E which are also referred to as events.   $\Omega$ is a set of subsets of E.  We refer to elements of $\Omega$ as events.  For the purpose of this chapter it is enough to think of $\Omega$ as the set of all possible subsets of E.  A random variable is simply a function from E to the real numbers, $X : E \rightarrow R$.
We define the distribution function associated with a given random variable X, $F : R \rightarrow [0, 1]$ by $F(x) = P(\{\omega \in E | X(\omega) \leq x \}) = P(X \leq x)$.  Thus, the value of the distribution function is a function from R, the real numbers, to numbers between 0 and 1 such that the value of F at x, $F(x)$, is the probability that the random variable will get a value that is less than or equal to x.   The following example and exercises clarify these definitions.

\begin{example}
Consider a fair dice.   In that case, $E = \{1, 2, 3, 4, 5, 6\}$ which are the possible results of a throw of the dice and we consider each possible result to be equally likely.  Thus, $P(i) = \frac{1}{6}, i = 1, \dots, 6$. As $E$ is finite, all possible subsets of $E$ are possible events in the probability space.  Thus, $\Omega = P(E)$.  A possible event is the event of getting some even number when throwing the dice.  It is represented as $EVEN = \{2, 4, 6 \}$ and is indeed a subset of E.  In addition, $P(EVEN) = P( \{2, 4, 6 \}) = \frac{3}{6} = \frac{1}{2}$.  Let us define a random variable $X(i) = i, i = 1, \ldots 6$.  We then have $F(3) = P(\{\omega \in E | X(\omega) \leq 3 \}) = P(\{1, 2, 3\}) = \frac{3}{6} = \frac{1}{2}$.  
\end{example}

\begin{exercise}
\label{Chapter:MLtesting:Excercise:distribution}
Calculate $F(i)$ for $i \in \{-1, 0, 1, 2, 5, 6, 10, 10.1\}$.
\end{exercise}

As part of our experiment that tests the ML based system we focus on some random variable and estimate different aggregates of the random variable behaviour such as its average.  We would like to do that in a robust manner without assuming anything about the distribution.  As an example, in the case of a chatbot ML based system we would like to determine the average number of service calls that will be routed to a human agent. We would like to be able to say that on the average, say, 20$\%$ of the calls will be routed to a human agent.  In addition, we would like to say that with high confidence the average will fall between 18$\%$ to 22$\%$ of the calls.  In order to do that we would need to estimate the distribution of the random variable. We next study the concept of the empirical distribution that will let us estimate the distribution of the random variable.   

We are given a sample, $S = \{x_1, \ldots, x_n\}$ of the random variable, chosen randomly and independently from $E$.  Next, we define the empirical distribution function, $F_e(x) = \Sigma_{i \in S} \frac{I(x_i \leq x )}{n}$ where $I(condition)= 1$ if the condition is true and $0$ otherwise. Intuitively, $F_e(x)$ estimates the probability that the random variable will have a value that is less than x, i.e., $P(X \leq x)$, by determining the percentage of the elements of the sample S that are less than x.  The next example helps clarify the concept.

\begin{example}
Consider the fair dice again. Define the random variable to be $X(i) = 1$ if and only if $i$ is even and zero otherwise.   Assume we got the following sample 
$S = \{ 0, 0, 1, 1, 1, 0, 1, 1, 0\}$.   Then $F(0) = P(\{1, 3, 5\}) = \frac{1}{2}$ as $X(1) = X(3) = X(5) = 0 \leq 0$ but $F_e(0) = \Sigma_{i \in S} \frac{I(x_i \leq 0 )}{n} = \frac{4}{9}$.  Importantly note that the empirical distribution is not equal to the distribution of the random variable.  
\end{example}

The following exorcise develops the intuition of when the empirical distribution represents the distribution of the random variable.  In fact, as the size of the sample increases the empirical distribution converges to the distribution of the random variable.  

\begin{exercise}
\label{Chapter:MLtesting:Excercise:emericalDistribution}
Write a simulation that obtains a large S from a fair dice distribution.   What is $F_e(0)$ converging to?   Same question for $F_e(1) - F_e(0)$?

Solution can be found in Appendix \ref{chapter:solution_appendix} in \ref{Chapter:MLtesting:Excercise:emericalDistribution_sol}
\end{exercise}

TBC - add material on average and variance here or in the appendix by expanding \ref{probability}. 

Recall that the Bernoulli distribution, $Br(p)$, is obtained by a trail that has success probability $p$. Success is denoted by $1$ and failure by $0$. For each point, $x_i$, in the sample $S$, $P(x_i \leq x) = F(x)= P(I(x_i \leq x ) = 1)$.   Thus, the random variable $I(x_i \leq x)$ is distributed Bernoulli with probability $F(x)$.   Or $I(x_i \leq x)  ~~_{\tilde{}}~~Br(F(x))$. We thus have $E(I(x_i \leq x)) = F(x)$ and $V(I(x_i \leq x)) = F(x)(1-F(x))$.

Let's pause here to better appreciate what have just happened.  We are taking a \gls{non parametric} approach.   One of the fundamental patterns of \gls{non parametric} statistics has just been realized above.  The argument above applies for any distribution $F()$.   Regardless, we have made an observation that connected the statistics of interest, in this case, $I(x_i \leq x)$, with some known distribution,$Br(F(x))$, thus returning to the well known grounds of parametric statistics!. \footnote{Eitan - clarify - add some definitions and examples to distinguish between parametric and non parametric statistics.}

\begin{exercise}
Using the fact that $I(x_i \leq x)$ is distributed Bernoulli prove that $E(I(x_i \leq x) = F(x)$ and $V(I(x_i \leq x)) = F(x)(1-F(x))$. As a consequence what is the average and variance of $F_e()$?  
\end{exercise}

As a consequence $E(F_e(x)) = E(\Sigma_{i \in S} \frac{I(x_i \leq x )}{n}) = \frac{\Sigma_{i \in S} E(I(x_i \leq x )}{n} = \frac{n F(x)}{n} = F(x)$.   $F_e(x)$ is thus an unbiased estimator of $F(x)$.  In addition, $V(F_e(x)) = V(\Sigma_{i \in S} \frac{I(x_i \leq x )}{n}) = \frac{\Sigma_{i \in S} V(I(x_i \leq x )}{n^2} = \frac{nF(x)(1-F(x))}{n^2} = \frac{F(x)(1-F(x))}{n}$.   The important point to note here is that as $n$ grows the variance of $F_e(x)$ vanishes making $F_e(x)$ an excellent estimate of $F(x)$. 

Let's take a deep dive into the behaviour of the empirical distribution.  

\begin{definition}
We are given a sequence of random variables $X_1, X_2,\ldots$ with a corresponding distribution functions $F_1(), F_2(),\ldots$.   In addition, we are given a random variable $X$ with distribution function $F()$.  We say that $X_1,X_2,\ldots$ converges in distribution to $X$ at $x \in R$ if $F_n(x) \rightarrow F(x)$.  If that is the case for each $x \in R $ for which $F()$ is continuous we say that $X_1, X_2,\ldots$ converges to $X$ in distribution.   
\end{definition}

\begin{exercise}
\label{excercise:525_gn_fn}
Consider the series of functions $f_n(x) = \frac{1}{2}(1-\frac{1}{n})^2x+(1-\frac{1}{n})$, $n \in N$.   Show that for each $n \in N$, the function $g_n(x) = f_n(x)$ if $-\frac{1}{(\frac{1}{2}(1-\frac{1}{n}))} \le x \le 0$ and $0$ otherwise is a density function.  Also show that for each $x \in R$,  the series $f_n(x)$ converges as $n$ goes to infinity to $f(x) = \frac{1}{2}x+1$.   Show that $g(x) = f(x)$ if $-2 \le x \le 0$ and $0$ otherwise is also a density function.  Is it also true that the random variables defined by $g_n(x) = \frac{1}{2}(1-\frac{1}{n})^2x+(1-\frac{1}{n})$ converges in distribution to $g(x)=\frac{1}{2}x+1$?   If this is the case what needs to be proven?

\href{https://colab.research.google.com/github/GuysBarash/MLBook/blob/main/ex_5_2_4.ipynb}{Click here for a simulation} or see appendix \ref{chapter:solution_appendix} in solution \ref{excercise:525_gn_fn_sol}

\end{exercise}

\section{Control interval example with unlimited sampling with replacement}


We provide an example of obtaining a \gls{control interval} for a \gls{ML model}.  We assume unlimited access to labeled data that represents data at deployment time.   That is an ideal assumption. Much of our discussion in latter sections will focus on how to remove that assumption using \gls{bootstrapping} and Monte Carlo techniques but the 
\gls{non parametric} statistical approach will remain the same.   As previously mentioned, We will be applying mostly non parametric statistics throughout this chapter as we typically do not know the type of probability distribution.  This is another complication that is typical to modern ML work.  An exception to this rule is when we are dealing with averages and large samples.   The central limit theorem will then guaranty that the distribution of the average converges to the normal distribution and we'll be able to use that to utilize parametric techniques.

We assume a model $f()$  was developed that given an image $x$ determines if $x$ is a dog or a cat.  As we are given access to unlimited set of labeled data of cats and dogs that represents the data we will encounter when the model is deployed, we sample from the data and get a fresh set of labeled cats and dogs 
$(x_i, y_i), i=1, \ldots, n$. $x_i$ are the images and $y_i$ is the label $y_i \in \{cat, dog\}$.  We then calculated the sampled accuracy $a = \frac{\sum I(f(x_i) = y_i)}{n}$ where $I(c)$ is the indicator function that is equal to $1$ if the condition, $c$, is true and zero otherwise. 

We make the following observation. If we repeat the above procedure, and obtain a fresh random sample, we will get different results for the accuracy.  Thus, accuracy is a statistic that estimates the real accuracy of $f()$.  In order to determine the actual accuracy with high confidence we need to develop a control interval.   We proceed as follows.  We repeat the above procedure $k$ times and obtain $k$ accuracy scores $a_1, \ldots, a_k$.  We order them in descending order, wlog, $a_1 \ge a_2 \ge , \ldots , \ge a_k$.  We take away $2.5\%$ of the first numbers in the list and $2.5\%$ at the end.  The remaining numbers define the control interval.   The accuracy will lie in this interval in probability $95\%$.  

\begin{example}
We make the connection to the empirical distribution studied in the previous section explicit through an example.  Assume $k = 1000$ and $a_i = i$.  Thus, $2.5\%$ of the top will be the $25$ numbers $1000,\ldots,976$ and the bottom $2.5\%$ will be the the $25$ numbers $1,\ldots,25$. According to the definition of the empirical distribution, $P_e(a \leq 25) = \frac{25}{1000} = 2.5\%$, and $P_e(a \leq 975) = \frac{975}{1000} = 97.5\%$.   Thus, $P_e(a > 975) = 1-\frac{975}{1000} = 2.5\%$.  It turns out that $P_e( 25 < a \leq 975) = 95\%$.  As discussed in the previous section, when $k$ grows that will be a better and better approximation of the unknown $P()$ distribution.  
\end{example}\footnote{Eitan - check why $P_e$ is used here and not P.   Also give a continuous example.}

\begin{exercise}
Repeat the example with $k=100$ and $a_i = i$.  Be careful to handle the boundaries correctly.  
\end{exercise}

\begin{example}
\href{https://colab.research.google.com/drive/1CIA9_CHjoqAaLrrDwSdExffKcPtC6AmE?usp=sharing}{Click here for an example of a non parametric confidence interval.}  As the data is sampled from the normal standard distribution, and we take 2.2$\%$ of the largest and smallest points we expect the $96\%$ confidence interval to be approximately $[-2, 2]$ when the sample size increases.  
\end{example}

\section{Bootstrapping}

Bootstrapping is a way to overcome budget limitations.    Ideally we would like to obtain a set of fresh samples as previously explained and obtain a confidence interval for the accuracy or other performance measures of the ML model using non parametric statistics. As discussed this is then used to anticipate and control the performance of the system in the field.  Additional usage includes to compare models and determine if indeed one of them is better than the other or if the difference in their performance is a result of noise.  We thus see that the techniques we are discussing here influence the end to end process of development of a ML solution.  We deep dive into that the coming sections.  

In practice we can not obtain unlimited number of fresh data samples representing the field data.  In fact, if we are lucky, we have one sample that represents the field data and is labeled.  As long as this sample is not used for training the model we can use it to estimate the field behaviour of the model, but how do we overcome our budget limitation, namely, having only one sample?  We follow the bootstrapping procedure to obtain new data samples.   We call such data samples bootstrapped data samples.  Technically, we repeatably sample with replacement from the one sample that we have to get our new sample and then proceed as before.   
To understand bootstrapping consider a data sample of blue and red balls.  The probability of getting a blue ball is $\frac{1}{3}$ but we do not know that.  If the data sample is big enough we will have roughly $\frac{1}{3}$ of the balls being blue.   This is the big "if"  of bootstrapping - we require that the data sample distribution represents the real distribution.   Now if we sample balls randomly from the data set we will get a new data set that will also have the $\frac{1}{3}$ blue balls proportion. It is "as if" we have sampled a fresh sample where as we actually sampled from our existing sample.   This is the core idea behind bootstrapping.  We can do that as many times as we like to get a set of "fresh" samples and then apply the non parametric procedure described in the beginning of the chapter for the dogs and cats classification example to obtain a confidence interval on the accuracy. We treat the bootstrapped samples as fresh samples hence the name "bootstrapping".        

\begin{example}
\href{https://colab.research.google.com/drive/1XbuklUZrB5h6wUneAMJhboaIGCnSEZXV?usp=sharing}{Click here} to see an example of a bootstrapping and non bootstrapping confidence interval for the balls. The expected range should be around $\frac{1}{3}$.  Note that the sample size should be big enough in order for the bootstrap confidence interval to work.
\end{example}

\begin{example}
\href{https://colab.research.google.com/drive/1v6VudBYzgrymG-ZjQqMTWKlsvRm0GAUb?usp=sharing}{Click here for an example of trained models performance analyzed non-parametric confidence interval.}
\end{example}

\begin{exercise}
\label{mlTesting:dataBalance}
Bootstrapping can be also used when training a ML model.  In this exercise we consider how.  A training set, D, of 200 images of cats and 800 images of dogs is given. 
\begin{enumerate}
    \item If you sample with replacement from D 2000 times what is the average number of cat images that you are going to get?
    \item Can you change the sampling procedure so that the average number of cats will be $\frac{1}{2}$
    \item How will you use the second sampling method in the training of a ML model that classifies dogs and cats?
\end{enumerate}
\end{exercise}

\section{Using the central limit theorem instead}

Given our idealized assumptions of independent random sampling with replacement the central limit theorem could have been applied as well in order to obtain a confidence intervals.  In any case, it is a good idea to apply, if possible, more than one method to validate the confidence interval being developed.  

Recall that the normal variable is close under linear transformations.   Specifically if $X$ is normally distributed $N(\mu, \sigma)$ then $Y = aX+b$ is distributed with normal distribution $N(aX+b, |a|\sigma)$.

\begin{exercise}
Prove that 
if $X$ is normally distributed $N(\mu, \sigma)$ then $Y = aX+b$ is distributed with normal distribution $N(aX+b, |a|\sigma)$. Is that correct for $a = 0$?
\end{exercise}

The central limit theorem states that if $X_1, \ldots, X_n$ are sampled independently from an identical unknown distribution $F()$ that has a finite average $\mu$ and a finite standard deviation $\sigma^2$ then setting $S_n = \frac{\Sigma_{i= 1}^{i = n} X_i}{n}$ we have that $\frac{\sqrt{n}(S_n - \mu)}{\sigma}$ approaches the standard normal distribution, $N(0, 1)$, for sufficiently large $n$.  

Setting $a = \frac{\sigma}{\sqrt n}$ and $b = \mu$ we get that $\frac{\sigma}{\sqrt n}(\frac{\sqrt n(S_n - \mu)}{\sigma}) + \mu$  is distributed $N((\frac{\sigma}{\sqrt n})0+\mu, (\frac{\sigma}{\sqrt n})1 = N(\mu, \frac{\sigma}{\sqrt n})$. In other words, $S_n$ is distributed $N(\mu, \frac{\sigma}{\sqrt n })$. Thus, in order to develop a confidence interval we need to estimate the average and standard error of $S_n$ assuming it is normally distributed.  

TBC - explain the difference between the two possible estimates to the standard deviation as n is large it does not matter.  Give a Python example.

\begin{example}
\href{https://colab.research.google.com/drive/1L2n-9wsPbjq8_5kikkVRjfuC4078-276?usp=sharing}{Click here} for an example of how a large sample that is distributed normally can be used to estimate its average and standard deviation and thus obtain a confidence interval.  
\end{example}




\chapter{Drift Detection and ML Solution Retraining}
\label{Chapter:drift}

\newcommand{\nth}[1]{#1^{\text{th}}}

As previously mentioned, ML systems typically consist of input data and a target that is to be modeled or predicted on the basis of the input by an ML model. To the degree it is able, the ML model captures the relationship between the input and output target learned from training data.  The model will then be deployed (used to generate predictions) on another set of input data.  An underlying assumption of the exercise of building a model is that the input and target data in deployment will be similar to that in learning; if not, the model's predictions may not be trustworthy, and its performance (e.g., accuracy) will be unstable and different than on the training data.  What we call `drift' will be changes in the underlying data, whether or not they cause the model performance to change.  

This chapter is organized as follows.  We will first introduce basic notions of the types of drift and their effects on the model.  Then, we will discuss ways of detecting that drift has occurred, as well as some techniques to overcome these challenges.  Lastly, we will mention some statistical issues that arise in drift detection.  Since drift detection is a very wide topic, this is not exhaustive but will rather cover some basic concepts.

\section{Types of drift
\label{ssec:types_of_drift}}

Let $\mathbf{X}$ denote input data.  That is, $\mathbf{X}=\begin{bmatrix}\mathbf{X}_1,\mathbf{X}_2,\dots,\mathbf{X}_d\end{bmatrix}$, meaning each of its $d\geq 1$ columns\footnote{$\mathbf{X}$ is univariate if $d=1$ and multivariate if $d\geq 2$.} is a vector $\mathbf{X}_i$ of $n$ observations of a random variable or feature\footnote{See Appendix~\ref{ssec:random_variables}.} $X_i$.  Let $y$ be another feature that is the target random variable we wish to predict by an ML model based on $\mathbf{X}$.  $\mathbf{X}$ can denote input data of any kind, such as embeddings of natural language and or image instances, or structured tabular data; $y$ can denote a class or numeric-valued feature, in which case the ML model is a classifier or regressor. 

Letting $p(\cdot)$ denote an arbitrary probability distribution\footnote{See Appendix~\ref{ssec:random_variables}.}, let $p(y\mid\mathbf{X})$ denote the true, unobserved, probabilistic relationship\footnote{The notation `$a\mid b$' denotes $a$ conditioned on, or determined by $b$;  See Appendix~\ref{ssec:joint_distributions}}~between the value of the predictor features $\mathbf{X}$ and the target $y$; we aim to model $p$ by, say, some classifier model.  If a given dataset $D$ consists of $\mathbf{X}$ and $y$, the joint distribution of data observations in it can be denoted $p(y,\:\mathbf{X})$. This distribution can be decomposed\footnote{This follows from the laws of decomposing joint probability distributions into products of conditional distributions.  See Appendix~\ref{ssec:joint_distributions}}~as $p(y,\:\mathbf{X})=p(y\mid\mathbf{X})\times p(\mathbf{X})$, since the observations are assumed randomly sampled from an underlying population distribution $p(\mathbf{X})$.

\begin{figure}[h!]
\centering
\includegraphics[scale=0.5]{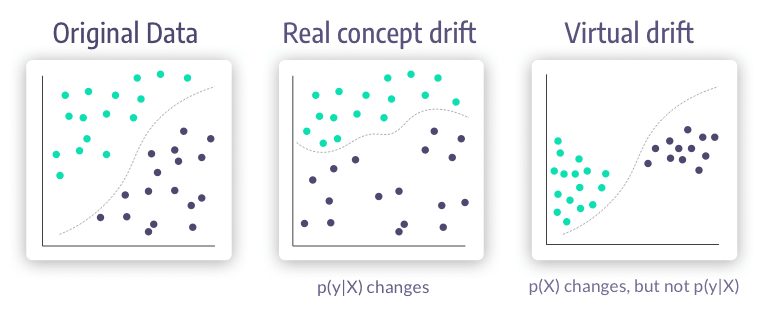}
\caption{Examples of types of drift (Source: \cite{Iguazio_drift}).}
\label{fig:boundary_drift}
\end{figure}

Mathematically, drift between two observed datasets $D$ and $D'$, can be expressed as saying that $p(y,\:\mathbf{X})\ne p'(y,\:\mathbf{X})$, that is, their respective joint distributions  differ.  If we further decompose both as above, if there is drift, then either $p(y\mid\mathbf{X})\ne p'(y\mid\mathbf{X})\ne$, or $p(\mathbf{X})\ne p'(\mathbf{X})$, or both.  Figure~\ref{fig:boundary_drift} illustrates these for the case where $\mathbf{X}$ consists of two numeric features (the horizontal and vertical coordinates), displayed as a scatterplot of points; for each point $x\in \mathbf{X}$, its binary target class $y$ is indicated by point being either hollow or filled.  In this simple case, the classes $y$ are completely separable in that the curved line is able to completely separate the points by color (class value).  The line thus abstracts $p(y\mid \mathbf{X})$ in that, say, $p(y=\textrm{blue}\mid x)=1$ if $x$ is on one side of the curve and 0 if on the other.

In Figure~\ref{fig:boundary_drift}, the left image shows the baseline $p(y,\:\mathbf{X})$, with the other two showing different drifted $p'(y,\:\mathbf{X})$.  For illustration, imagine that the two features plotted are AGE (horizontal, say from age 16 to 65 years) and CREDIT RATING (vertical, where higher values indicate better credit), each point corresponds to an applicant for a student loan; the class $y$ is the bank's decision, indicated by the color of the point, with black being `reject' and green being `approve'.  Here we see that at younger ages, any credit rating will get you a loan (green), while as age increases, only the applicants with better credit will be approved, with the approval threshold increasing.

In the center image is what is typically called `\textbf{concept drift}', where the relationship $p(y\mid\mathbf{X})$, illustrated by the placement of the class decision boundary, has changed. In this case, it means that, say, a young applicant with a lower credit rating ($x$ in lower left corner) would now be rejected when he previously would be accepted, since this point would now fall `under' the decision boundary. In this image, $p(\mathbf{X})$ has changed relative to the left one, since the points are located in different places, but $p(\mathbf{X})$ could have also remained constant.  The right image shows an example of `\textbf{virtual drift}', where only the sampling distribution $p(\mathbf{X})$ (point placement) has changed to $p'(\mathbf{X})$.  The decision boundary, however, remains the same as in the left image, meaning that a given input $x\in\mathbf{X}$ would still have the same class, since $p(y\mid\mathbf{X})$ is constant. Drift has occurred in that $p'(y,\:\mathbf{X})\ne p(y,\:\mathbf{X})$, but only through changes in $p(\mathbf{X})$.

The intuition of distribution drift in Figure~\ref{fig:boundary_drift} can be extended to any input $\mathbf{X}$, whether higher-dimensional or non-numeric inputs like images, sound recordings, etc., however in these cases the distribution $p$ may be difficult to characterize mathematically, and high-dimensional points may be difficult to visualize.  Also, $y$ can be a numeric target, like income in dollars, rather than a class, or consist of multiple target values per observation (multi-output).

We note that there is disagreement among practitioners as to how to precisely define terms for various types of drift. \cite{MRACH12} represents one attempt to formulate coherent definitions.  The joint $p(y,\:\mathbf{X})$ can alternatively be decomposed as $p(y)\times p(\mathbf{X}\mid y)$, changing the order of conditioning from that above.  In \cite{ADFRZ2020}, for instance, $\mathbf{X}$ represents a classifier's measured confidence on an input image instance of class $y$, and only $\mathbf{X}$ and $y$ are observed, not the intermediate images, and drift in $p(y,\:\mathbf{X})$ is detected.  Here, the drift is induced by changing $p(y)$ (introducing a previously unseen class) rather than $p(\mathbf{X}\mid y)$ (characteristics of handwriting of each digit).

\section{Measuring distribution differences
\label{sec:measuring_distribution_differences}}

In Section~\ref{ssec:types_of_drift}, we introduced drift as
indicating $p(y,\:\mathbf{X})\ne p'(y,\:\mathbf{X})$ when $p(\cdot)$ and $p'(\cdot)$ are determined on two different datasets (or samples) $D$ and $D'$.  To determine if there is drift, we want to measure degree of difference or distance between the two sample distributions. Depending on the objective, a drift analysis may be conducted on one or more of the following:
\begin{itemize}
    \item $p(y)\textrm{ vs }p'(y)$, the class label distributions
    \item $p(\mathbf{X})\textrm{ vs }p'(\mathbf{X})$ or $p(\mathbf{X}\mid y)\textrm{ vs }p'(\mathbf{X}\mid y),\textrm{ for each }y$; that is, the overall or class ($y$)-conditional data distribution
\end{itemize}

Furthermore, we note that each dataset $D$ is actually treated as a random sample draw from some theoretical unobserved `population' dataset $\mathcal{D}$ (see \cite{LSHGOZ2003}, page 20) with associated distribution $P(\cdot)$ (as contrasted with $p$ or $p'$). For instance, in the example mentioned in Figure~\ref{fig:boundary_drift}, $D$ and $D'$ may represent samples (say, of sizes 1,000 each) of loan applicants in the USA in the months of January and July, 2021, respectively.  The respective populations $\mathcal{D}$ and $\mathcal{D}'$, with distributions $P$ and $P'$ respectively, are all such applicants in the USA in these two months; the samples were obtained because, perhaps, data on all such applicants (population) is not easily attainable.  The sample distributions are considered estimates (denoted $\hat{\phantom{}}\:$) of the respective population distribution; thus, we can say $p=\widehat{P}$ and $p'=\widehat{P'}$, meaning, for instance ``$p\textrm{ is an estimate of }P$".

Due to random sampling, if two sample distributions $p$ and $p'$ are drawn from the same population distribution $P$, we would not expect that $p=p'$ exactly.  Instead, if we try to decide if drift were occurred, we want to avoid considering small differences that may be due to random sampling as `drift'.  Rather, we typically try to make inferences such as the following: given $p=\hat{P}$ and $p'=\hat{P}'$, how likely is it that $P\ne P'$?  That is, the drift question often tries to infer whether the two observed samples $p$ and $p'$ are different enough (more than by random chance) to indicate that the unobserved $P$ and $P'$ themselves are different.  If they are, this is considered `drift'.  This is the nature of statistical hypothesis testing (see \cite{LSHGOZ2003}, page 375).

Such statistical drift tests can be conducted if the distribution can be characterized mathematically.  Typically, such tests are called `two-sample tests' since they compare two random samples for equality, rather than, say, evaluating one sample to a fixed baseline value or distribution.  However, sometimes a mathematical representation of the distribution cannot be constructed, and thus a measure of distance between the two samples $D$ and $D'$ may be measured instead; \cite{SamBlog} (Q4) reviews some examples.  Here, we will just cite these measures without detailed elaboration.

\begin{exercise}
Assume that $\mathbf{X} = \{ x \in R^3 | x_i \ge 0, x_1+x_2+x_3 = 1 \} $ and that $y = \{-1, 1\}$.  Assume that the model m is accurate iff $x_2 \leq \frac{1}{2}$.  The model was trained with p being uniformly distributed on $\mathbf{X}$. What is the accuracy of the model on p?  Next, the distribution on $\mathbf{X}$ is shifted to be uniformly distributed on $\{x \in \mathbf{X} | x_2 \ge \frac{1}{2}\} \bigcup \{x \in \mathbf{X} | x_3 \ge \frac{1}{2}\}$.  What is the accuracy of the model on the new distribution?  What type of drift occurred in this case?  Suggest a change to the problem conditions so that the other type of drift will occur.  
\end{exercise}

\subsection{Two-sample distribution difference measures on \texorpdfstring{$\mathbf{X}$}{X}}
\label{ssec:two_sample_continuous}

For univariate samples ($\mathbf{X}$ consists of a single feature), many tests exist; for those we describe in the following, we also list the test function name as it can be found in the relevant Python package, such as \texttt{scipy} (\cite{2020SciPy-NMeth}).\footnote{The cited test names are accurate as of \texttt{scipy} version 1.8.0.}  

Given two samples $D$ and $D'$, sometimes one is interested to know if they have, say, the same (arithmetic) mean, median, or variance\footnote{The arithmetic mean (average) and median ($\nth{50}$ quantile) are measures of central tendency; see \cite{LSHGOZ2003}, pages 131--133.  The variance is a commonly-used measure of variability; see \cite{LSHGOZ2003}, page 144--151.}, rather than whether the whole distributions $P=P'$. In some cases, $P\ne P'$ while their means, medians, or variances can be unchanged.  However, a change in mean, median, or variance indicates the distributions differ as well, particularly in the aspect that is important to the user.  For instance, a teacher may want to know if the mean scores on a math test increased, regardless of whether the distribution changed (e.g., more students received very low or very high scores).

In these cases, the two-sample Student T-test (\texttt{scipy.stats.ttest\_ind}) is often used to detect differences in the mean; Bartlett's test (\texttt{scipy.stats.barlett}) or Levene's test (\texttt{scipy.stats.levene}, if the samples don't appear normally-distributed\footnote{Various tools can be used to assess whether a sample appears normally-distributed.A quantile-quantile (QQ) plot (e.g., \cite{LSHGOZ2003}, page 280) can visualize these differences.  There are also statistical tests to assess similiarity to a normal distribution, such as the Shapiro-Wilk (\texttt{scipy.stats.shapiro}), D'Agostino-Pearson (\texttt{scipy.stats.normaltest}), and Anderson-Darling (\texttt{scipy.stats.anderson}).} can detect differences in variance.  The Kruskal-Wallis (KW, \texttt{scipy.stats.kruskal}) tests can be used to test for equality of medians.  If the data $\mathbf{X}$ are binary-valued ($X$ can be coded as $\in\{0,1\}$), the means are proportions restricted to $[0,1]$; in these cases, Yates' difference-in-proportions test (\cite{Y1934}) can be used. 

\begin{exercise}
\label{excercise:622}
As an example for when summary of the data such as the sample average and sample variance are of interest consider the following situation.
\begin{enumerate}
\item 
Assume that a feature $X \in R$ in the data D is normally distributed with average 0 and standard deviation (i.e., the square root of the variance) $\epsilon > 0$.  $\epsilon$ is assumed to be very small.  What is your expectation of the impact of X on the dependent variable y? 
\item  Assume that X, T, and Z are independent variables.  Also assume that the relation we are trying to learn is $X+T+Z = y$.  Write a program that generates X, T, and Z independently each from a normal distribution.  Have the average of X, T and Z be 0.  In addition, the standard deviation of X is some small $\epsilon > 0$ while the standard deviation of T is 100 and the standard deviation of Z is 1000.  Generate n samples $(x_i, t_i, z_i)$.  For each generated triple $(x_i, t_i, z_i)$ calculate $x_i+y_i+z_i = y_i$.  Also calculate the sample average of $x = \frac{1}{n} \sum x_i$. calculate $x + y_i + z_i = y^{'}_i$.   What is the maximal difference between $y_i$ and 
$y^{'}_i$?  what does it tell you about the learning of y?  
\end{enumerate}

\href{https://colab.research.google.com/drive/1h6sHrpPEhcwglv_majEZ_1ousbYGnSXC?usp=sharing}{Click here for a simulation} or see appendix \ref{chapter:solution_appendix} in solution \ref{excercise:622_sol}

\end{exercise}

\begin{exercise}
Consider the Dirichlet distribution.  It is an important distribution used in Bayesian analysis but for the purpose of this exercise all we need to know about the distribution is that it is determined by $a_1, \ldots, a_k$ parameters, $a_i > 0$.  Whenever we sample a point from the distribution we will get a point $x = (x_1, \ldots, x_k)$ in the simplex, i.e., $x_i \ge 0$ and $\sum x_i = 1$.  The average value of the i$'$th coordinate is $\frac{a_i}{\sum a_i}$ while the variance for the i$'$th coordinate is $\frac{(\frac{a_i}{a})(1-(\frac{a_i}{a}))}{a+1}$ where $a = \sum a_i$. Given the average and the variance can you determine the parameters $a_i$ of the distribution?
\end{exercise}

\begin{exercise}
Give an example for two different distributions that have the same average and variance.  (Hint - pick up a uniform distribution with some average and variance and match it with a normal distribution that has the same average and variance. )
\end{exercise}

If one wants to determine the likelihood of $P=P'$, and not just equality of summary statistics, often nonparametric tests are used; these tests make minimal or no assumptions about the shape of the distributions.  Some common ones for continuous data are Kolmogorov-Smirnov (KS, \texttt{scipy.stats.ks\_2samp}),
Cram\'er-von Mises (CvM,\\ \texttt{scipy.stats.cramervonmises\_2samp}), or Anderson-Darling (AD,\\ \texttt{scipy.stats.anderson\_ksamp}); the AD test can test simultaneous distributional equality of $k>2$ samples, not just two.  If testing equality of the distributions of orderings of values (rather than the distributions over the values themselves), the Mann-Whitney test (MW,\\ \texttt{scipy.stats.mannwhitneyu}) is often used.  Say, for instance, each of $D$ and $D'$ is associated with a separate team, and the values are the times in minutes that each racer on the team finishes a race; assume that the winners are determined only by the relative order they finish and not by the absolute times.  In this case, the MW test indicates if the racers in the two teams are approximately evenly-matched in terms of the order (but not necessarily the measured times). 

The tests mentioned so far, like many others, rely on p-values, or significance levels (see \cite{LSHGOZ2003}, page 375) to make a decision.  The null hypothesis ($H_0$) in each case is specified as equality of either the distributions $P$ and $P'$ or the statistic of interest; a low p-value, below a pre-specified $\alpha\in(0,1)$, indicates that drift is likely because the distributions differ significantly, and thus the null $H_0$ should be rejected.  Because the statistical significance measured by p-values is known to have methodological issues, measures of \textbf{effect size} (\cite{LSHGOZ2003}, page 692; \cite{SF2012}), which more closely capture the magnitude of \textit{practical} difference between the two samples' summary statistics.  Some commonly-used measures are Cohen's $d$ and $h$ ($=\textrm{arcsin}(\sqrt{\pi_1}) -\textrm{arcsin}(\sqrt{\pi_2})$ for independent proportions $\pi_1$ and $\pi_2$) measures (\cite{C1988}) for testing two-sample differences in means and single binary proportions.  These measures are compared to specified thresholds rather than an $\alpha$ value, to make a significance decision.  For instance, for both the $d$ and $h$ measures, values of 0.2, 0.5, and 0.8 represent small, medium, and large differences, respectively (\cite{C1988}, pages 40, 198).    

The local kernel-density-difference test from \cite{D2013} (implemented in \texttt{RJournal of Modern Applied Statistical Methods} as the \texttt{ks} package, \cite{DWCG2018}) identifies regions where the two densities $p$ and $p'$ differ significantly, as opposed to simply \textit{if} they differ. An illustration on univariate data, with Python implementation, is shown in \cite{SamBlog} (Q9).

For multivariate samples, the distributions are often harder to characterize easily, except in certain parametric cases like multivariate-normal, without making distributional assumptions.  Wasserstein (also known as ``Earth-mover's") distance measures the distance between two distributions using transport theory, representing each observation as, say, a grain of sand that must be `moved' from $p$ to $p'$.  The more `distance' each `grain' must be moved to transform $p$ into $p'$, the greater the distance between them.  These metrics are implemented in Python as \texttt{scipy.stats.wasserstein\_distance} (univariate case) and \texttt{ot.emd} (multivariate).  Another nonparametric kernel-based distance is Maximum Mean Discrepancy (MMD, \cite{GBRSS2012}); this is illustrated in \cite{SamBlog} (Q4).

\subsection{Two-sample distribution difference measures on \texorpdfstring{$y$}{y}}
\label{ssec:two_sample_class}

In addition to the tests mentioned in Section~\ref{ssec:two_sample_continuous} on numeric-valued input data $X$, we can also test whether the distributions of observed labels $\mathbf{y}$ differ.  Again, this determination is often made by assuming unobserved population distributions $P$ and $P'$.  Because the values $\mathbf{y}$ are nominal-valued (categorical), such as the state (e.g., Alabama, Alaska,\dots,Wyoming) of residence of the loan applicant, $P$ typically takes the form of a multinomial distribution (\cite{LSHGOZ2003}, page 208).  The multinomial distribution $\mathcal{M}$ is denoted $\mathcal{M}(\boldsymbol{\pi},\:N)$, where $N\in\{1,2,\dots\}$ is a positive integer representing the sample size (number of observations), and $\boldsymbol{\pi}=\begin{bmatrix}\pi_1 & \dots & \pi_k\end{bmatrix}$ is a $k$-length vector where $0\leq \pi_i\leq 1,\:\forall\:i=1,\dots,k$, and $\sum_{i=1}^k\pi_i=1$.  Without lack of generality, let the $k$ potential label values be $\boldsymbol{\ell}=\{\ell_1,\dots,\ell_k\}$.  $\mathcal{M}(\boldsymbol{\pi},\:N)$ models $N$ label values, drawn independently, where each draw takes the $\nth{i}$ label $\ell_i$ with probability $\pi_i$.  Since all that is modeled is the total count of each label out of $N$ draws, and not the order, a vector $\mathbf{x}=\begin{bmatrix}x_1,\dots,x_k\end{bmatrix}$ can be modeled as a draw from $\mathcal{M}(\boldsymbol{\pi},\:N)$, if each $x_i$ is a non-negative integer and $\sum_{i=1}^kx_i=N$.  That is, $x_i$ is the observed occurrence, out if $N$, of label $\ell_i$.  Therefore, given two observed label vectors $y$ and $y'$, where the corresponding counts vectors are $\mathbf{x}$ and $\mathbf{x}'$ (assuming both have the same size $N$), we can determine if they appear to come from the same multinomial distribution $P$.

Probably the most-used test for equality of categorical counts is the 
one-way chi-square goodness-of-fit (\texttt{scipy.stats.chisquare}), not to be confused with the chi-square test for independence performed on a contingency table.  This test requires the sums of $\mathbf{x}$ and $\mathbf{x}'$ to equal ($N$ is the same).  However, it is a one-way test, since one of $\mathbf{x}$ or $\mathbf{x}'$ must be set as the `expected' and the other as the `observed' values, which affects the calculation; in contrast, in the two-way tests, there is no `baseline', and the order of specification of samples does not matter.  This test returns a p-value used in the drift decision.  

An alternative is to use the effect size metric Cohen's $w=\sqrt{\sum_{i=1}^k \frac{(\hat{\pi}'_i - \hat{\pi}_i)^2}{\pi_i}}$ (\cite{C1988}), where $\hat{\pi}_i=\frac{x_i}{\sum_j x_j}$ and $\hat{\pi}'_i=\frac{x'_i}{\sum_j x'_j}$ are the corresponding proportions in the `expected' ($\mathbf{x}$) and `observed' ($\mathbf{x}'$) samples, respectively. Cohen's $w$ is very similar to the chi-square statistic except that it operates on the proportions rather than the frequencies, and thus the samples sizes do not have to be equal.  The rule-of-thumb thresholds for $w$ are 0.01, 0.2, 0.5, 0.8, 1.2, and 2.0 for very small, small, medium, large, very large, and huge effect sizes (\cite{C1988}, page 227; \cite{S2009}).

The chi-square and Cohen's $w$ metrics are asymmetric tests in that one set of values is set as the `expected' ones. Note also that $w$ is undefined if any $x_i=0$, which makes it unusable if a category level in the observed $\mathbf{x}'$ does not occur in the expected $\mathbf{x}$.  Several alternative metrics exist.  The dissimilarity index (\cite{G1914}) is defined as $\hat{\Delta}=0.5\sum_{i=1}^k|\hat{\pi}_i - \hat{\pi}'_i|$, and a rule of thumb is that $\hat{\Delta}<0.03$ indicates the two distributions are very close (\cite{A2002}, page 329). Hellinger's distance is defined as $H=\sqrt{1-B}$, where $B=\sum_{i=1}^k\sqrt{\hat{\pi}_i\hat{\pi}'_i}$ is the Bhattacharyya coefficient (\cite{B1946}).  It has the property that $0\leq H\leq 1$, with $H=0$ indicating distributional equality; in addition, it obeys the triangle inequality.  Both $\hat{\Delta}$ and $H$ are symmetric.  Note that $\hat{\Delta}$ will reflect labels where, say, $\hat{\pi}'_i\ne 0$ but $\hat{\pi}'_0=0$ (label $i$ occurs in the observed but not expected distribution), but $H$ will not, since the two probabilities are multiplied.

Another metric is Jensen-Shannon distance (\cite{ES2003}), which is the square root of the Jensen-Shannon divergence.  This distance is implemented as \texttt{jensenshannon} in \texttt{scipy.spatial.distance} (\cite{2020SciPy-NMeth}).  It is also symmetric and obeys the triangle inequality.  Given the two probability vectors $\hat{\boldsymbol{\pi}}$ and $\hat{\boldsymbol{\pi}}'$, let $\boldsymbol{m}$ be their point-wise mean for each discrete value $i$.  The distance is defined as 
$JSD(\hat{\boldsymbol{\pi}}, \hat{\boldsymbol{\pi}}')=\displaystyle{\sqrt{\frac{D(\hat{\boldsymbol{\pi}}\parallel\boldsymbol{m}) + D(\hat{\boldsymbol{\pi}}'\parallel\boldsymbol{m})}{2}}}$, where $D$ is the Kullback-Leibler (KL) divergence.  This distance has the property that $JSD\in [0, \log_b{(2)}]$, where $b$ is the logarithm base used in $D$; the maximum bound is 1 if $b=2$ and $\ln{(2)}$ for natural logarithms.  KL divergence for probability vectors $P$ and $Q$, which is not symmetric, is defined as $D_{\textrm{KL}}(P\parallel Q)=\sum_i P_i\ln{\left(\frac{P_i}{Q_i}\right)}$.  The summation is done for $i$ for which both vectors are non-zero; thus, $\boldsymbol{m}>0$ whenever at least one is non-zero, so the equation above is defined.  The value $i$ at which $P_i\ln{\left(\frac{P_i}{m_i}\right)} + Q_i\ln{\left(\frac{Q_i}{m_i}\right)}$ is maximum is the level which differs the most between the two distributions.

\section{Drift in characterizations of data
\label{sec:drift_in_characterizations}}

In Section~\ref{sec:measuring_distribution_differences}, we presented a series of two-sample statistical tests, distance, and effect-size measures to detect distributional drift between two datasets $D$ and $D'$.  There, we assumed we could, or desired to, model the dataset or label distributions directly.  However, sometimes it is desired to model intermediate `aspects' of the data, or a model's performance on it, and detect drift on these aspects instead.  We present several examples here.

\subsection{Drift in data slices \label{ssec:drift_in_data_slices}}

Letting $D$ and $D'$ be tabular structured datasets of numeric or categorical features.  \cite{ARZ20} introduces the notion of a data `slice' rule as a conjunction of subsets of a group of feature; a subset on a feature is an interval with a minimum and maximum value if the feature is numeric, or a set of potential values if it's nominal.  For instance, an example of a slice $S$ is $\{\$30,000\:\leq\:\textrm{INCOME}\:\leq\:\$60,000\}\:\&\:\{\textrm{STATE}\in\{\textrm{New York},\textrm{Ohio},\textrm{Michigan}\}\:\&\:\{\textrm{SEX}\in\{\textrm{Male}\}\}$, which is defined on the three features INCOME, STATE, and SEX.  An observation falls in a slice if its feature values satisfy all conditions in the slice; for instance, any males with an income in the range \$30,000--\$60,000 and living in either New York, Ohio, or Michigan.  The size, or support of a slice on a dataset of $N$ total observations is the number of observations falling in the slice; the fractional support is the support divided by $N$.  Given a classifier model that returns predictions on $D$, \cite{ARZ20} present an algorithm to find a set of such slices where specifically the classification error rate of the model is higher than the average over $D$; such slices are called `error-based slices'.

Given two similar datasets $D$ and $D'$ and the classifier's predictions on each, \cite{ADFRZ2021} presents a method for detecting drift between $D$ and $D'$.  The drift is not detected directly on the datasets' feature distributions $p(\mathbf{X},\:y)$ and $p'(\mathbf{X},\:y)$, but rather by extracting a set of error-based slice rules on $D$, and detecting differences between this set and the rules when mapped to $D'$.  Note that the $K$ slices can overlap in that an observation can fall into more than one of them.  If $K$ slice rules are extracted, $\hat{\pi}_{1,i}$ and $\hat{\pi}_{2,i}$ be the observed fractional support of the $\nth{i}$ slice out of $K$ on datasets $D$ and $D'$, respectively.  If there is no drift between $D$ and $D'$, we expect $\hat{\pi}_{1,i}\approx\hat{\pi}_{2,i}$; the same is true of all $K$ slices.  A difference-in-proportions test (\cite{Y1934}, mentioned above in Section~\ref{ssec:two_sample_continuous}) can be used to test this; since $K$ hypotheses, one for each slice, are conducted, we have $K$ p-values $p_1,\dots,p_K$.  

Ultimately, we want a single decision of drift over all the slice p-values, and not just to see if each slice individually has drifted (changed in size).  This is done by using an adjustment for multiple comparisons (see \cite{LSHGOZ2003} page 424), specifically Holm's method (\cite{H1979}), which produces a single p-value with a statistical guarantee on the familywise error rate (FWER) of the decision (i.e., across the `family' of multiple hypotheses).    

In this example, \cite{ADFRZ2021} used an `indirect' drift test based on the set of slices, which has several advantages. First, it allows repurposing an existing technology for slice-extraction (\cite{ARZ20}) for drift detection; this method can be used in other settings where `useful' observation subsets can be defined.  Second, since the purpose of the slices was to locate concentrations of model errors by the feature values, the technique allows detection of drift specifically in these areas, which are particularly useful because they indicate likely changes in model accuracy, assuming stability of the slice rules, and not just drift in the feature distributions.  Thirdly, and most importantly, distilling the dataset into the simpler aspect of slice rules, which are now modeled by the univariate measure of proportion, simplifies the analysis from the multivariate case of actually modeling all the feature distributions.

\subsection{Drift in density-based slices}

\cite{AFRZZ2021} present a way to partition a dataset $D$ into a set of slices, which are defined in the same way as in \cite{ARZ20}. However, here the $K$ slices extracted differ from \cite{ARZ20} in two ways.  First, they form a partition, in that together they contain all observations in $D$ and that each observation belongs to exactly one of them (as opposed to \cite{ARZ20}, where there may be overlaps and observations that do not fall in any of the slices).  Second, these slices are constructed to contain observations with similar `spatial' density within the feature space, rather than classification error; furthermore, some slices may defined feature subsets that are empty, containing no observations.  

As discussed in Section~\ref{ssec:drift_in_data_slices}, this method distills the feature space of $D$, which may be high-dimensional, into the lower-dimensional abstraction of useful slice rules.  Furthermore, as in \cite{AFRZZ2021}, though not discussed there, these slices can be used for drift detection in a similar way by testing differences in univariate measures, such as their fractional support or volume.  

For instance, say the $K$ density-based slices on $D$ can be categorized as either `not very sparse' (type A), `very sparse' (type B), or `empty' (type C); the sparsity threshold can be decided based on, say, the average core distance, or slice volume divided by fractional support (see \cite{AFRZZ2021} for details).  In $D$, observations must fall only into either slices of type A or B, since those of type C are by definition empty.  Now, let each observation in another dataset $D'$ be mapped to one of the $K$ slices.  Then, each observation in $D'$ can fall into one of either types A, B, or C.  In addition, there is a new type, D, meaning that an observation contains at least one feature on which a slice rule is defined, whose value is outside of the ranges observed in $D$; for instance, a higher income than the highest observed in $D$, or a country of origin not observed in $D$.  Similarly to in \cite{ADFRZ2021}, if there is no drift in feature distributions between $D$ and $D'$, we may expect each type A, B, C, or D to have similar fractional support on the two datasets.  Significant drift in these proportions---after an adjustment for multiple hypotheses, or by the distance metrics in Section~\ref{ssec:two_sample_class}---may indicate drift.  Furthermore, this drift is easily explainable, in that we can point to which type (A, B, C, or D) changed the most, and identify some of the anomalous observations.  

\subsection{Drift in feature polynomial relations}

Given a tabular dataset $D$ of only numeric features, \cite{RARF2021} present a method to extract strong polynomial relations from it.  Say that the input data $\mathbf{X}$ contains $m$ features, denoted $X_1,\dots,X_m$.  A polynomial relation is an polynomial `equation' between, say, $X_1$ and some $k$ (e.g., 2) of the features, allowing feature interactions up to a limited degree $\ell$ (say, 2).  For example, $X_1\approx2 + 3X_2 - 5X_2X_3 + 1.5X_2^2$; let us denote this relation $\mathcal{L}_1$. The relation between $X_1$ and the expression involving $X_2$ and $X_3$ is found using linear regression; hence the $\approx$ symbol means that this is not a strict equality, but that there is some error, which is measured by the error term of the linear regression equation.  Strong relations are polynomials where the regression coefficient of determination $R^2$ is high, indicating a strong linear correlation between the true value (e.g., $X_1$) and the `prediction' of the polynomial on the other features $X_2$ and $X_3$.

Given a polynomial $\mathcal{L}_1$ determined on $D$, we can see how good its fit is on the same features in another dataset $D'$; in the absence of drift, we expect the fit to be about the same.  The degree of change in fit is quantified by the Bayes Factor of $\mathcal{L}_1$ on $D$ vs $D'$.  In experiments with simulated drift insertion in \cite{RARF2021}, it is shown that for relations that had high initial $R^2$ (strong), the Bayes Factor was more responsive to drift insertion than weaker relations; that is, the strong relations were better indirect sensors of feature drift.  Since many relations can be extracted from $D$, a correction should be done to adjust for the multiple comparisons performed.

As in the previous examples discussed above, drift analysis based on the relations, rather than $\mathbf{X}$ itself, can be simpler.  \cite{RARF2021} suggest constraining the relations by $k=\ell=2$, to prevent over-fitting and so they are relatively human-interpretable.  The logic of drift detection here is that a strong relation is likely a fixed aspect of the data that should be stable; for instance, we may expect a person's SALARY to have a fixed polynomial relationship to their years of EXPERIENCE and EDUCATION, and if this changes, it may indicate an underlying feature drift.  In particular, if an ML model is to be be deployed on $D'$, its performance may differ significantly from on $D$ if there is drift, if the model exploits existing feature correlations for its predictions.  Even if the intermediate relations themselves are not of interest, they can still be used as `detectors' of drift.

\section{Sequential drift detection}

So far, the statistical tests and examples of indirect drift detection (Section~\ref{sec:drift_in_characterizations}) dealt with a single decision on $D$ vs $D'$.  In many cases, however, we may observe an ordered sequence $D_1,D_2,\dots$, often ordered in time, and want to determine if any differ from the initial $D$.  The setting may be `online', meaning we can store and directly use the cumulative data $\{D_1,\dots,D_t\}$ that have arrived up to time index $t$, or not.  Here, we will ignore such distinctions.  

\subsection{Types of sequential detectors
\label{ssec:types_sequential}}
Previously, we considered the observed data to consist of two samples $D$ and $D'$.  In sequential detection, the data may consist of batches or samples, each known to come from a potentially different distribution $D_i$, or may just consist of a sequence of individual observations without any such demarcation.  In general, consider the data as a sequence $x_1,x_2,\dots$; $x_i$ may be multivariate, but we will initially assume it is univariate.  Here, the task often consists of identifying one or more time indices $t_1, t_2, \dots$  (typically called changepoints), if they exist, where the data distribution has changed, or drifted.  Due to the vast literature on sequential changepoint detection, we will not attempt to provide an in-depth review, but rather summarize some of the aspects of such algorithms.  

Broadly speaking, sequential algorithms can be thought of as either supervised, semi-supervised, or unsupervised.  In general, a supervised model, such as a classifier, returns a prediction (e.g., of the label), which is then compared to a ground truth value.  A supervised sequential detection algorithm can be used when, for instance, $x_i$ are observations for which the ground truth label $y_i$ is available\footnote{We mean that the ground truth is known (e.g., we know, whether the customer with attributes $x_i$ ends up buying the product ($y_i=1$) or not ($y_i=0$), from an observed interaction.  In contrast, say $x_i$ are mammogram images or hardware test logs, which require an expert to examine and assign a true label $y_i$ for.  It may be `expensive' (in terms of time, effort, and money) to determine $y_i$, and perhaps it may be feasibe to only label a small subset of instances.  The time aspect is particularly important: obtaining $y_i$ may cost no effort or money, but only time, in that it will be known only after a long time delay of the `arrival' of the initial data $x_i$.  For instance, $y_i$ is the success/failure of a treatment or death/survival of a patient with attributes $x_i$ in a clinical trial.  In these cases, it may be difficult to assess the model acuracy for changes, since $y_i$ may not be available for comparison with the prediction $\hat{y}_i$; alternative approaches which we discuss, such as monitoring changes in $x_i$ or in the model's prediction confidence, may be used.}, and for which an ML model has been trained to predict a label $\hat{y}_i$.  Typically, these detectors measure changes in a sequence of model accuracy measures, such as $I(\hat{y}_i=y_i)$ (binary indicator of correctness, from which the accuracy rate can be estimated) or the error $\hat{y}_i-y_i$ when the target is numeric rather than a class.  The relevant drift aspect is typically whether the error (in terms of its distribution or average) changes over time; as before, this may be separate from whether the distribution of $\{x_i\}$ themselves have changed.

In unsupervised\footnote{Clustering is an example of a unsupervised algorithm.} algorithms, there may or may not be an associated ground truth $y_i$.  Unsupervised drift detection algorithms typically try to detect drift in the distribution of $\{x_i\}$ themselves.  In the simplest cases, $x_i$ are univariate and numeric, and the sequence is monitored for changes in the average value or variance.  A comprehensive review and taxonomy of unsupervised detection algorithms is given in \cite{GCGD2020}. If $x_i$ is itself a binary variable, then one can monitor changes in it by applying some of the same supervised detectors, which typically assume the observed binary variable is $I(\hat{y}_i=y_i)$.  

\chapter{Optimal Integration of the ML Solution in the Business Decision Process}
\label{optimal}

As previously discussed ML embedded systems are non deterministic.  They make mistakes by design.  What we hope to achieve instead of a bug free system is a system for which the error on the choices we want to utilize in the business decision process is statistically controlled.  Such control enables the appropriate allocation of human resources for the correction of errors made by the ML embedded system and the insurance that overall we obtain a stable process that increases the profit of the organization.  In addition, the way the ML embedded system is used may need to be updated over time as the factors impacting the performance of the ML embedded system may change.  The last issue, drift identification, was dealt with in chapter \ref{Chapter:drift} while the question of statistical control was studied in chapter \ref{Chapter:MLtesting}.    

The ML embedded system is integrated in a business process and should help increase value obtained from the process by the organization.   For example, there are many consideration involved in reaching a decision of whether or not to give a loan to a bank's customer.  One of the consideration is if the loan payments to the bank will occur on time.  Assume that we have developed a ML embedded system that predicts if the loan will be paid on time by a given customer. We want to integrate the prediction of the system in the overall decision process followed by the bank.  The embedding should be made in such a way that the bank will maximize the profit from giving loans to customers.  Giving loans that will be paid on time and not giving loans that will not be paid on time can help maximize the overall bank profit but this is not the only consideration.  Other considerations may apply and influence the final bank decision.   For example, if the bank can get the payments for the loans by other means such as accessing the customers assets.  Let's spell out two fundamentally different ways in which we can integrate the ML system in the business process.
\begin{enumerate}
    \item If we can trust the recommendation of the ML system we can just act according to the recommendation.  For example, a software problem is reported.   The software is composed of components.  Each component has an owner that can solve problems of the component that she owns.  If we can trust the ML system to route the software problems correctly most of the time we can just route them automatically according to the ML recommendation.   As long as the routing recommendation made by the ML system is mostly correct the value from the business process is clear as we no longer require a team to route the software problems to the appropriate component.
    \item The ML system recommendation is correct most of the time but can not be "blindly" trusted.  In our problem routing scenario a person is required to consider each ML recommendation.   This can still be beneficial as given the recommendations, especially if the ML system provides the reasons for the recommendation, the time required to route each software problem is reduced. 
    \item A hybrid case.   There are some conditions in which we have established, through an appropriate experiment, that we can trust the ML system "blindly" and some conditions for which we need a person to inspect the recommendation and make the final decision.  A simple example of that is if we have established that for some software components we can route the problem report automatically according to the ML system recommendation and for some components we can not and a person should make the final decision.
\end{enumerate}

If the ML recommendation is not followed blindly (cases two and three above), a person sometimes should make the final decision taking into account the ML recommendation or potentially ignoring it all together.  In this situation whether or not the person makes a correct decision depends on the incentive defined by the organization.  Thus, as part of the optimal business definition process, if there are decision points in which a person needs to make a decision while taking into account a ML system recommendation that can not be fully trusted, optimal incentives should be defined for that person.  This calls for the application of mechanism design, see \href{https://www.investopedia.com/terms/m/mechanism-design-theory.asp}{link}.   We will attempt to get a better understanding on how this can be achieved in practice in the next section.

\section{Business process optimization and incentive design}

We start with an example of optimization of the business process associated with the ticket routing ML embedded system.  The ML embedded system uses several models and utilizes different parts of the problem report such as the report text and the report meta data.  In addition, the system may utilize other sources of information such as the software installation configuration.   Finally, the system may also use some deterministic rules based on certain error codes.  For example, if out of memory error is reported route the problem to the memory management of the software.   As discussed in chapter \ref{unitSystem}, the ML system is thus typically composed of a hybrid decision tree.  We are emphasising this as one may think the assumptions we are making next only apply to a single model and they typically do not.   Hence the consistent use of the terminology ML embedded systems throughout this work.  

We assume that the software has four components, namely, $c_1, c_2, c_3, c_4$.   Previous experiments have shown that the average conditional probability of correct routing decision given that the decision is one of the components,$c_1, c_2, c_3, c_4$. is  $P(correct|c_1) = 0.6, P(correct|c_2) = 0.7, P(correct|c_3) = 0.8, P(correct|c_4) = 0.9$ respectively.  A human resource is available that can analyze the problem and correctly route it at the cost of 1 unit.   We are also given that the observed probabilities of reporting a problem by the ML system from each of the components $c_1, c_2, c_3, c_4$ is $\frac{1}{4}$.  The average performance of the system if the human resource is not used is thus
$\sum_{i = 1, 2, 3, 4}\frac{1}{4}P(correct|c_i) = 0.75$.  We consider applying the human resource only in case that the system predicts $c_1$ which we refer to as the elimination of $c_1$ policy.   On the average that will be a quarter of the time, so on the average we expect to pay 250 units of payments in a 1000 software problems.   What is the expected performance of the joint business process that utilizes the human resource in that way?   It is expected to be completely correct for $c_1$ thus $P(correct|c_1) = 1$.  We will have the average performance of $\sum_{i = 1, 2, 3, 4}\frac{1}{4}P(error|c_i) = 0.85$.  We thus increased the average performance of the system from $0.75$ to $0.85$ at a cost of 250 units of cost in a 1000 software reported problems.  This highlights the trade-off involved in applying the human resource.  Consider the following exercise to better understand the concept. 

\begin{exercise}
Assume the routing problem as explained above.  
\begin{enumerate}
    \item What will be the average performance if we only eliminate policy $c_2$?  Same question for policy $c_3$ and $c_4$?  What are the average costs in that case?
    \item We randomly choose to apply the human resource in probability $p$. What will be the average performance of the system and the average cost in this case?
    \item You are given a budget of 100 cost units per a 1000 software problems how would you spend it?  What will be the average performance then?
    \item What is the standard error of an elimination policy $c_i$ given that we know that the standard errors of each of the conditional probability $P(correct|c_i)$ from previous experiments?  How would you check the stability of an elimination policy?
    \item What will be the impact of change in the probability of the ML system reporting a problem in a components.  For example, what will be the impact of the ML system reporting a $c_1$ problem in probability $0.7$ and $c_2, c_3, c_4$ in probability $0.1$ each?  Assume the conditional probabilities of the ML system being correct given the component prediction stays the same.  Explain why such a phenomenon is possible?
    \item What will be your recommendation if you do not have an estimate of the probability in which the ML system will choose one of the components? Explain.
\end{enumerate}
\end{exercise}

We can thus consider a general category of policies, namely elimination of $c_i$ given that the ML system recommended routing to $c_i$ in probability $p_i$.  The expected accuracy in such a case will be 
$\sum_{i = 1, 2, 3, 4} \frac{1}{4}(p_i + (1-p_i)P(correct|c_i))$.  We denote this expected accuracy by $P(correct|p_1,\ldots,p_4)$.  We can think of our problem as the problem of maximizing $P(correct|p_1,\ldots,p_4)$ under a given budget con train. We are ensured that this optimization problem has a solution.  See next exercise for details. 

\begin{exercise}
Show that $P(correct|p_1,\ldots,p_4)$ is continuous and that the set of possible utilization of the budget constrains, e.g., if the budget is 250 cost units per a 1000 software problems and is utilized so that $\frac{1}{4}(p_1+p_2+p_3+p_4)1000 = 250$, is compact.  Deduce that the optimization problem has a a maximum.  See appendix for details on why this is the case.     
\end{exercise}

In a way the above approach to the problem is the simplest possible approach as either the ML embedded system is making the routing decision or the human resource does.  Concerns are thus "separated".   We can measure the performance of the ML embedded system when its decision is trusted and separately measure the performance of the human resource when she makes the decision on the routing.  In other words, for each routing decision there is a single "owner", either the ML embedded system or the human resource.  Next we consider a more hybrid scenario in which the two are brought together to make the final decision.

Assume that the human resource spends up to some time bound on the routing decision and that the accuracy is proportional to the time spent.  For example, assume that the human resource spends up to 2 hours on the routing decision and that the accuracy in routing of a decision that was made after t time is $\frac{t}{2}$.  The cost now of applying the human resource is the overall time that the human resource spent on making routing decisions.  We also assume that other assumption on the routing problem and the ML embedded system remain the same.  Assume that the overall time the user can allocate for a 1000 software problems is T. The human resource can perfectly handle $n = \frac{T}{2}$ routing decisions (ignore the reminder).  One approach could be to assume that we have n cost units as before and solve the previous optimization problem.  

\begin{exercise}
Assume the routing problem described above.  Further assume that we would like to raise the average accuracy from $0.85$ to $0.9$.  One way to achieve that would be to spend $0.9 \times 2$ time on each routing decision for which the ML embedded system recommended $c_1, c_2, c_3$.  This will raise the the conditions probabilities $P(correct|c_i), i=1, 2, 3$ to $0.9$.  How big should T be to implement this approach?
\end{exercise}

We now consider the problem of incentive.   We observe that there are two different type of decision makers.  The first design the overall business process.   For example, decides what type of elimination policy to choose in the problem routing example.   The second decision are the decision made by a human resource that participate in the decision process.   In our routing problem example the human resource making the routing decision.  The first principle of incentive definition is to tie incentive to the 
the part of the organization business goal controlled by the decision maker.  In our ticket routing example the decision maker that optimizes the entire business incentive should be tied to the entire business process performance.  For example, in the ticket routing example the business process optimizer should be rewarded for a decrease in the overall time required to solve problems, a decrease in the overall expense required to solve the software problems, and so on.  In contrast, the human resource used to route software problems only controls the correct and efficient routing of a given software problem.  Thus, she needs to be rewarded in proportion to volume of correct routing and in reverse proportion to time it took her to make the decision.  Note to provide such incentives the organization will need to measure the performance of the system.   This should be thus part of the system requirement and experiment design from day one of the ML embedded system development!   

\begin{exercise}
Consider the bank loaning example.  Assume that two ML embedded system are developed.  The first estimates if a loan will be return on time by a customer with accuracy 0.8.  The other ML embedded system estimate if a customer will increase her business with the bank as a result of given a loan.  There are budget con trained human resources that can make the two decisions.  The bank is expanding so if the business with the customer is likely to increase and the customer is likely to pay the loan the bank policy is to given it.  Define appropriate optimization process and incentives for the business process optimizer and human resource decision maker.  Make additional assumption as needed similar to the one made in previous exercises of this chapter.  
\end{exercise}

\chapter{A Detailed Chatbot Example}
\label{industrial}



\section{Chatbots as a ML embdded system}
Chatbots are becoming a key channel for customer engagement. Chatbots are automated systems through which users can interact with the business through a natural language interface. For many customers, the chatbot provides their first interaction with the business and serves as the ‘face’ they meet—and their first impression. 
Automation that can create a positive and rich customer experience , and enable repeat business, must be able to ‘understand’ the customer as a human would and respond accordingly.  However, many chatbots fail to provide a high-quality customer experience because they do not understand the customer’s intent, are not designed to cover enough situations, or even fail to respond appropriately to the user request. 
To provide the best possible customer experience, the chatbot has to be reliable, be consistent, interpret user intents correctly, and respond appropriately. The chatbot has to respond by comprehending the underlying intent behind the users’ utterances.  This is something that can only be ensured through comprehensive training and testing that is geared specifically to the chatbot’s business performance, its conversational responses, and interactions

\begin{figure}
    \centering
    \includegraphics{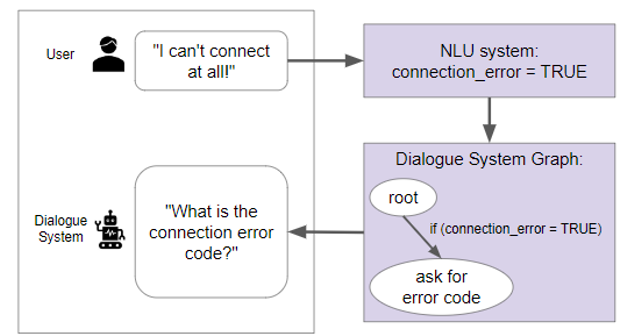}
    \caption{\label{fig:chatbot_arch_exmaple} An example of a chatbot interaction with one turn, in which the user complains they cannot log in. The system first needs to understand this user utterance, and here a classifier is employed to identify the intent of the user. Then, a rule-based engine, or an execution graph, actually decides on the response that will be given. }
\end{figure}

Chatbot technology usually comprises two basic  components, as shown in Figure~\ref{fig:chatbot_arch_exmaple}: 
\begin{enumerate}
    \item \label{enum:classify}	A machine learning (ML) natural language processing (NLP) based intent classifier that can process what the user is saying, and 
    \item \label{enum:flow} A conversation flow orchestrator that incorporates domain knowledge and is driven by the business actions and content extracted from past human-to-human dialogs and company documents.  Typically the orchestrator is rule base and does not apply use ML.
\end{enumerate}

This chatbot architecture may be represented as a non deterministic tree.  The ML models that classify the user intents (item \ref{enum:classify} above)  typically occupy the upper levels of the non deterministic decision tree. The rule-based flow orchestrator (item \ref{enum:flow} above) typically occupies the rest of the tree and is deterministic. 

\begin{exercise}
An online shop sells home repair equipment for householders and professionals.  A model, $m_t$, was developed that identifies if a customer is a preferential or a householder.  Another model, $m_a$, determines if the customer would like assistance in the use of some equipment. Additional models, $m_h, m_p$, were developed that give recommendations for additional purchase for householders and professionals respectively. Finally, a model $m_c$ was developed that determines if the customer would like to complain about an equipment that was purchased. There is a rule based system, $r(m_t, m_a, m_c)$, that handles the customer requests.   Design a non deterministic decision tree that utilizes the above ML models and the rule based system.  Determine the expected accuracy of each path in the decision assuming the accuracy of the trained ML models are given.
\end{exercise}

\section{Quality challenges}

Testing a chatbot requires assessing not only the quality of the classifier but also the end-to-end conversation including the subsequent intermediary system actions (business functions) that complete the conversational interaction. This testing must be carried out in the pre-deployment stage, before the chatbot is deemed production worthy.  Of course, the testing is also needed once the chatbot is in production, to check for functional consistency and monitor for continuous improvement.  As with any ML solution, it will need periodic evaluation and testing to flag recalibration and the need for adjustments.  

The big challenge of chatbot development lies in getting enough quality data to train the chatbot and test it thoroughly in the first place. This is more problematic if the chatbot has yet to interact with customers, meaning there is no history of interactions and conversations to use as a test bed. The trainer or the tester of the system must be able to provide enough sentences that can predict what the users will say at runtime. This is challenging. Usually very little data is available to test or even train the chatbot. There exist powerful data augmentation technologies that can help overcome this challenge. 

Depending on the level of business logic one may want to develop, building a simple question-answering bot can be pretty straightforward. But, a more serious investment of resources is needed to enable the bot to deal with more complicated user queries.  When clients, and customers of clients, are exposed to an early or low-quality version of a chatbot, their perception and satisfaction will be affected by its quality. 

The nature of human language makes it impossible for software tests to cover all possible situations.  Although websites and smartphone apps use predefined interactions based on common user interface components like buttons, hyperlinks, or text-entry, the integrated chatbots have to cover both the directed or expected and the unexpected or free format conversational variations. This is where the importance of comprehensive testing comes in.

Testing should begin early in the solution life-cycle. This is fundamentally different from the complementary activities of analyzing the system once it is deployed  for continuous improvement and retraining. The data that is available is much more limited in the testing phase, and does not include real user interactions. For example, actual contextual variables for input that needs to be identified during the flow are mostly provided when real conversations exist.  The majority of data available in the testing phase is related to the machine learning task of classifying those first important user utterances into the correct category of intent. This is the training and/or test data for the intent classifier. To provide reasonable coverage, this data must be extensive and representative of real user interactions with the chatbot.

While conversation flow modeling also needs to be developed early on, it is based primarily on the input of the person developing the virtual assistant. Once the virtual assistant has been released to users, there exist actual conversation logs. These logs can then be analyzed to understand which conversations were abandoned and why. This data also provides an opportunity to analyze conversations that went wrong and improve the conversation flow and the intent classifier based on user interaction. 

It is important to develop approaches to predict such un-handled conversation flows based on the data available during testing. The idea is to provide the system with examples of challenging utterances so the chatbot developer can improve its design and implementation even before the first release.  

\section{Control for chatbot success}
TODO <<Ateret lets do it together>

Monitoring completion status during operation is needed in order to measure the actual chatbot system business value.
This is different from monitoring the intent classifier accuracy. One can can view this as system level monitoring vs. unit testing.

\section{Trend analysis}

When a chatbot is released to users there is data of conversations and their completion status. Un-handled logs, or traces from those human-chatbot interactions that did not succeed, are of special interest.

In addition to improving the chatbot training data and potentially also design, un-handled logs may be an indication of actual changes, trends or drift. These changes may be due to changes in the environment or in the topics. There is a need to understand when changes are intermittent and should actually best be ignored, and when they indicate change of trend or drift and should be accounted for, for example by retraining the ML models.

Covid-19 questions provide an example. At the beginning of the pandemic most questions to Covid-19 related chatbots were around asking for information about the pandemic. Then as governments published restrictions, questions changed to be around the nature of the restrictions. When vaccines were developed, questions changed yet again to be around the vaccines and related recommendations and warnings. 

The reader is referred to Chapter \ref{Chapter:drift} for a comprehensive discussion of drift and its identification and ML models. 
TBC Eitan - remove include files that where part of the original book template
\appendix
\label{App:AppendixA}
\chapter{Mathematical Background}

This appendix covers mathematical and statistical background which facilitate the understanding of the main text and ML in general. 

\section{Optimization}

Optimization is at the heart of ML. Typically ML algorithms solve some optimization problem.  We will give examples of such optimization problems in other sections of this appendix.   

We first consider a general optimization setting and describe a pattern of numerical optimization algorithm that will apply to most of our optimization problems.  

Assume $x_n \rightarrow x$ in $R^n$ and that $x_n = A^{n-1}(x_1), n=2,\ldots$.   Further assume that $A: R^n \rightarrow R^n$ is continuous.  The target of our optimization is some $O \subset R^n$.
Assume we have some function $Z : R^n \rightarrow R$ such that $Z(A(x)) < Z(x), x \in R^n - O$ and $Z(A(x)) \leq Z(x)$ for $x \in O$ then we have the following.

\begin{lemma}
Under the above assumptions $x \in O$
\end{lemma}
\begin{proof}
$x_n \rightarrow x$ therefore $A(x_n)$ which is $x_2, x_3, \ldots$ also converges to $x$.  But as $A()$ is continuous $A(x_n)$ also converges to $A(x)$.  Thus, $A(x) = x$.   As a result it is not true that $Z(A(x)) < Z(x)$ therefore $x \in O$. 
\end{proof}

$Z()$ is a descending function that guide the search for an element in $O$. in practice it could be the function itself for which we are seeking a minimum but it could also be the and other monotonic descending function.  Another choice we will use many times is a direction in which the gradient of the function is descending.    

\begin{figure}[h!]
\centering
\includegraphics[scale=0.5]{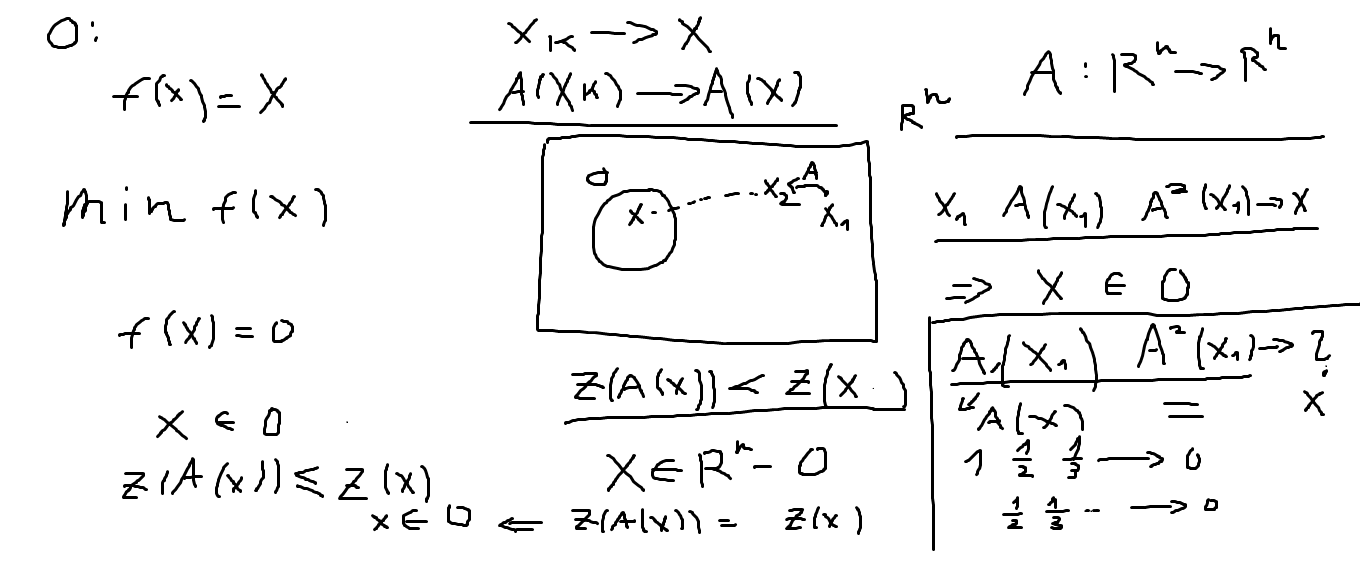}
\caption{When does a descending sequence converges to the optimum?}
\label{fig:universe}
\end{figure}

\begin{exercise}
In the above proof we use the fact that removing a finite number of elements from a converging sequence does not change its limit. We study this below.
\begin{enumerate}
    \item $1, \frac{1}{2}, \frac{1}{3}, \ldots \rightarrow 0$ but what about $\frac{1}{2}, \frac{1}{3}, \ldots$. Explain why. 
    \item $a_n \rightarrow a$ means that for all $\epsilon > 0$ there exist a $N_{\epsilon}$ so that if $n > N_{\epsilon}$ then $|a_n - a| < \epsilon$.  Prove that if $a_n \rightarrow a$ so that the sequence $b_1 = a_k, b_2 = a_{k+1}, \ldots$ for any $k > 0$.
    \item Can you modify the above proof to work for the removal of any finite set of elements from a converging sequence? That is if $a_n \rightarrow a$ and a finite set of elements is removed from $a_n$ to obtain $b_n$ then $b_n \rightarrow b$.
\end{enumerate}

\end{exercise}

\begin{exercise}
Consider the following process.  Given $f(x) = x^2$ we would like to find the root $f(0) = 0$ by starting from some pair of points $(x, -x), x > 0$.   We define the following process $a1 = (x, -x)$.  Given that $a_n = (a_n^1, a_n^2)$ if $(\frac{a_n^1+a_n^2}{2})^2 > 0$ then $a_{n+1} =(\frac{a_n^1+a_n^2}{2}, a_n^2)$ else $a_{n+1} =(a_n^1, \frac{a_n^1+a_n^2}{2})$.
\begin{enumerate}
    \item Define $A()$ and $Z()$ in this case.
    \item What happens to $Z()$ at $(0, 0)$.   What happens to $Z()$ at $(a, b)$ for which $a>0, b<0$
    \item Prove that for any $n$ $a_n^1 = -a_n^2$.
    \item Is $A()$ continuous?
    \item Write a program that implements the above numeric algorithm and check that it works (finds the root of $f(0) = 0$).  Generalize for any $f()$.
\end{enumerate}
\end{exercise}

\section{Probability}
\label{probability}

Here, we describe and show notation for probability concepts used in this book.

\subsection{Random Variables and Distributions}
\label{ssec:random_variables}

Random variables are introduced in Section~\ref{randomVariable}.  Here, we explain the concept in a less-mathematical way.  A variable (often denoted by a capital letter, e.g., $W,X,Y,Z$, etc.) or `feature' is a an object that can take a value.  For instance, $W$ may take any value in $\{1,2,3,4,5,6\}$, $X$ any real number between $[0,1]$, $Y$ any value in  $\{\splitatcommas{\textrm{Afghanistan},\textrm{Albania},\dots,\textrm{Zimbabwe}}\}$, and $Z$ any non-negative integer $\{0,1,2,3\dots\}$.  Here, $W$, $X$, and $Z$ would be considered numeric, and $Y$ would be categorical (see e.g., \cite{LSHGOZ2003}, page 28).  In addition, $W$ and $Z$ are discrete, and $X$ is continuous. 
Random variables are particularly useful when they represent real measurements.  For instance, $W$ may represent the result of a die being rolled, $X$ can be any potential probability value (which must be between 0 and 1), and $Y$ is a person's native country, and $Z$ can be the answer to the question `how many children do you have' (none or any positive integer).

A random variable also has a distribution associated with it, which determines the probability or likelihood of the variable taking each of the potential values.  For instance, letting $P$ denote `probability' we can speak of $P(Y=\textrm{Germany})=0.1$ or $P(0.1\leq X\leq 0.134)=0.2$, for example. Each of these specific or range of values is known as an `event'.  The probabilities are often calculated from a (random) sample of values of the variable (e.g., the native countries of a random sample of individuals; see \cite{LSHGOZ2003}, page 20--25).  Specific values of the variable are often denoted with lower-case letters (e.g., $w,x,y,z$).  So, for instance, we may consider $P(Z=z)$ for each of $z\in\{0,1,2,3\}$.

Distributions are often denoted $p$ or generality; thus we can use the shorthand $p(X)$ and $p(Y)$ to denote the distributions of $X$ and $Y$.  Distributions of numeric variables often have shapes that can be described by summary statistics (see \cite{LSHGOZ2003}, pages 131--153) such as mean, median, variance, kurtosis, etc.  Categorical variable distributions can be be visualized by bar  charts (see \cite{LSHGOZ2003}, page 101).

\subsection{Joint Distributions}
\label{ssec:joint_distributions}

The joint distribution of two or more variables is the probability of both variables taking specific values simultaneously.  Here, we will only illustrate joint distributions for two variables.  For instance, using the shorthand above, $p(X,\:Y)$ is the joint distribution of $X$ and $Y$.  For instance, letting both $X$ and $Y$ be categorical variables, where $X=\begin{cases}1, &\text{person speaks Spanish}\\0 &
\text{otherwise}\end{cases}$ \qquad and $Y$ being their native country, $p(X, Y)=P(X=x,Y=y)$ is the joint distribution.  Thus $P(X=1,Y=\textrm{Spain})$ is the probability (in a given sample or population) that a person both is from Spain and speaks Spanish.  This probability is zero if there are either no people from Spain, no Spanish speakers, or no people who fulfill both conditions (even though there may be, say, Spanish speakers who are not from Spain).

If both $X$ and $Y$ are numeric, a scatter plot can be used to visualize the joint distribution (see \cite{LSHGOZ2003}, pages 165--169).  Otherwise, except for particular cases (e.g., multivariate normal), the distribution is hard to characterize in a closed-form expression. 

A conditional probability (see \cite{LSHGOZ2003}, pages 194--197) is denoted $P(X=x\mid Y=y)$, which likewise yields a conditional distribution $p(X\mid Y)$.  The variable conditioned on ($Y$) is to the right of the vertical bar, and the variable whose probability is calculated given the conditional ($X$), is on its left.  This means that to calculate the probability, the sample is first restricted to values of the conditioning variable ($Y$) satisfying a condition.  For instance, consider $P(X=1\mid Y=\textrm{Spain})$.  Thus, we first consider only people whose native country is Spain; among these, we calculate the probability that a person speaks Spanish.  This probability is likely nearly 1 because there is a strong association between these two features (nearly all people from Spain speak Spanish), which is very different from the un-conditional joint probability above.

We note that a joint distribution can be decomposed into a product of a conditional and a univariate distribution.  For any variables $X$ and $Y$, $p(X,Y)=p(X\mid Y)p(Y)$.  Since the ordering of variables in the joint is arbitrary, we can also say that $p(X,Y)=p(Y,X)=p(Y\mid X)p(X)$.  That is, for instance, to determine the joint probability $P(X=1, Y=\textrm{Spain})$ we could first calculate $P(Y=\textrm{Spain})$ (likely small), then calculate the conditional probability of speaking Spanish given that one is from Spain (nearly 1).  Or, we could first calculate the probability $P(X=1)$, then calculate the probability of being from Spain (as opposed to, say, Argentina, Mexico, or any other country) among Spanish speakers.  Either way is equivalent.

We noted above that $X$ and $Y$ in this example have a strong association; that is, a person from Spain is more likely than a random person to speak Spanish, and a Spanish speaker is more likely than a random person to be from Spain.  Such an association whereby knowing, say, the value of $Y$ impacts the probability of $X$ being a particular value is called dependence, and lack thereof is called independence.

\subsection{Independence}

\begin{definition}
Two events A and B are independent if $P(A \cap B) = P(A)P(B)$.
\end{definition} 

We next define independence of two random variables.  
\begin{definition}
Two random variables X, and Y are independent if any pair of events $A_X$ and $B_Y$ defined by X and Y respectively are independent.
\end{definition}

\begin{example}
If X defines the height and Y the weight of people then an event defined by X may be $A_X = \{180 cm \le X \le 190 cm\}$ and an event defined by Y may be $B_Y = \{70~ kg~\le X \le 80~kg\}$.  Saying that X and Y are independent will mean that $P(A_X \cap B_Y) = P(A_X)P(B_Y)$.  The same forum la will hold for any two events $A_x$ and $B_Y$ you can define by X and Y.
\end{example}

\begin{exercise}
Are X and Y in the example above intuitively independent?  Explain your answer. 
\end{exercise}

When the two variables are independent we have that the average of their products is equal the product of their average, i.e., $E(XY) = E(X)E(Y)$.  We prove this for the case in which both variables have a finite set of values.  A similar proof holds when the number of possible values is countable (infinite but can be described as a series) or if the random variable is continuous.  In the latter and integral is used instead of a sum to obtain the proof.

Assume X may obtain values $x_1, \ldots x_n$ with positive probability and Y may obtain values $y_1, \ldots, y_n$ with positive probability.  In that case by definition $E(XY) = \sum_i \sum_j x_i y_j P(X = x_i \cap Y = y_j)$
Using the independence of X and Y we get $ = \sum_i \sum_j x_i y_j P(X = x_i)(Y = y_j)$
In the inner sum $x_i$ and $P(X = x_i)$ are constant thus we get $ = \sum_i  x_i P(X = x_i) \sum_j y_j  (Y = y_j)$
next we identify the average of Y as the inner sum thus
$ = \sum_i  x_i P(X = x_i) E(Y)$
but now $E(Y)$ is a constant so we get 
$E(Y) \sum_i  x_i P(X = x_i)$
which is clearly equal to
$= E(X)E(Y)$.  This completes the proof that $E(XY) = E(X)E(Y)$.

\subsection{Controlling the number of defects in a production line}
\label{parametricControl}

Consider a production line that produces a series of identical items.  Each item can be either defective or not.  We want to control the percentage of defected items.  We model the situation parametrically using a series of Bernoulli variables $X_i$ with probability of having a defect p.  For a batch of n items the production lines produced, inspection may yield k defected items.  We want to set our expectation about the production line. Fixing a small probability $\alpha$ we want to understand when seeing k or more defects has probability $\alpha$.   In other words, we would like to determine when the probability of the number of defects being greater than k for a batch of n items is $\alpha$ under the assumption that the probability of seeing a defected item is p.   If we then see more than k defects in a batch of n items we will say that the interpretation of the production line as a system that produces defect in probability p is no longer reasonable as we are seeing an unlikely number of detects (bigger than k).  We will take some action such as stooping the production line and re-collaborating its parameters.  In such a case we will say that the production line is "out of control".

We next formally capture the above reasoning. We require that $P(t \ge k) = \alpha$.   As the probability of t defects is $\binom{n}{t}p^t(1-p)^{n-t}$ we get that the probability of seeing more than k defects is $\sum_{t \ge k}\binom{n}{t}p^t(1-p)^{n-t}$.  We thus are interested in finding k so that $\sum_{t \ge k}\binom{n}{t}p^t(1-p)^{n-t} = \alpha$.  

\begin{exercise}
\label{excercise:a22}
Write a code that finds k. Play with different values of n, k, and $\alpha$ to gain an intuition on the value of k.  Is there a computational limitation to the calculation of k when n is large?

\href{https://colab.research.google.com/drive/10-2OR1zBRtrjNr52WvvU3r2MDEbLbpog?usp=sharing}{Click here for solution} or see solution \ref{excercise:a22_sol} in  appendix \ref{chapter:solution_appendix}.

\end{exercise}

\section{Bayesian Networks}

Recall that any joint distribution $P(X_1, \dots, X_n)$ of the random variables $X_1, \ldots, X_n$ can be factored in the following way $P(X_1, \dots, X_n) =  P(X_n|Z_1, \ldots, X_{n-1}) \times \ldots \times P(X_2|X_1) \times P(X_1)$.  The order of the factoring is arbitrary chosen thus any order may be applied.

\begin{example}
\label{BN1}
Assume the following four Boolean random variables: $R$ for whether or not it rained, $SH$ for whether or not the car was washed, $W$ for whether or not the floor is wet and $S$ for whether or not someone has slipped. The following factorization of the joint distribution applies $P(R, SH, W, S) = P(R|SH,W, S)\times P(SH | W, S)\times P(W|S) \times P(S)$.  Intuitively, this means that for any possible values of the variables say $R = true, SH = true, W = true, S = false$ we can calculate the probability $P(R = true, SH = true, W = true, S = false)$ by calculating $P(R = true |SH = ture,W = true, S = false)\times P(SH = true | W = true, S = false)\times P(W = true|S = false) \times P(S = false)$ or calculating the probability that we did not slip, multiplied by the probability that the floor is wet given that we did not slip, and so on. 
\end{example}

\begin{definition}
We say that a random variable X is independent of a random variable Y if $P(X, Y) = P(X) \times P(Y)$.  As $P(X, Y) = P(X) \times P(Y|X)$ we also get that in such a case $P(Y | X) = P(Y)$.   Similarly $P(X | Y) = P(X)$.
\end{definition}

A Bayesian network (BN) captures the independence and dependence between random variables.  One simple view of a BN network is the corresponding factorization of the joint probability of random variables $V = \{X_1, \ldots, X_n\}$ that it represents.   

\begin{definition}
Assuming that we are given a DAG with vertices $V = \{X_1, \ldots, X_n\}$ and a set of edges E.  We will say that $(V, E)$ is a BN if $P(X_1, \ldots, X_n) = \prod_{i} P(X_i| parents(X_i)) $ where $parent(X_i) = \{X_j | (X_j, X_i) \in E\}$.   
\end{definition}

\begin{example}
\label{BN2}
Revisit example \ref{BN1} above.  Set $V = \{R, SH, W, S\}$ and $E = \{(R, W), (SH, W), (W, S)\}$.  Interpreting (V, E) as a BN we obtain that $P(R, W, SH, S) = P(S | W) \times P(W | R, SH) \times P(SH) \times P(R)$ which is intuitively an appealing factorization.     
\end{example}

\begin{exercise}
what are the Independence assumption Incorporated in the BN representation of example \ref{BN2}.  Hint - write down an appropriate gunnel factorization and use independence assumptions to achieved the factorization represented by the BN.  
\end{exercise}

\begin{lemma}
We are given a BN with three variables, X, Y, and Z and edges $(X, Z), (Y, Z)$. We next show that in such a case X and Y are independent.
\end{lemma}
\begin{proof}
$P(X, Y) = \sum_{z} P(X, Y, Z) = \sum_{z} P(z | X, Y)\times P(Y)\times P(X) =  P(X) \times P(Y) \times \sum_{z} P(z | X, Y) = P(X) \times P(Y)$.   The last inequality is valid as $\sum_{z} P(z | X, Y) = 1$, i.e, $P(Z | X, Y)$ is a distribution probability over the values of Z given some specific values of X and Y.
\end{proof}

\section{Decision Theory}

Chapter \ref{optimal} discusses the interface of the ML embedded system with the business process. As the ML embedded system is non deterministic, we are not certain if its output is correct or not. Dealing with such uncertainties is at the heart of decision theory which we introduce next.  

In decision theory a decision maker is given a set of alternatives or actions A she needs to choose from.  In addition, a possible set of states of the world, S, is given but the decision maker does not know which of the states in S holds.  Once the decision maker chooses an $a \in A$, and given that the unknown to the decision maker state of the world is $s \in S$, the decision has a loss $l(a, s)$.

What decisions would we consider to be rational?  A conservative approach to the problem should assume the worst; given that the decision maker takes the action a, we then check what is the maximal possible loss in this case.  In other words, we calculate $l(a) = max_{s \in S} l(a, s)$. Next, we determine for which action, $a \in A$, the worst possible loss, l(a), is minimal. We thus find the action, $a_0$, such that $a_0 = argmin_{a \in A}~l(a) = argmin_{a \in A}~max_{s \in S}~l(a, s)$.   This is sometimes called the $minmax$ principle and is conservative as in reality there is no adversary who is attempting to force the worst possible loss on the agent given the agent's action.

Another approach is to assume some distribution of the states S representing the likelihood of each state.  Given such an assumption the expected loss of the decision maker when taking an action a is  $l(a) = E(l(a, s))$.  It is then rational for the decision maker to choose an action that minimizes $l(a)$.  This is sometimes referred to as the Bayes rule.

\begin{example}
Consider a decision problem with two actions $A = \{a, b\}$.  we are given that if b is chosen the states of the world is determined and the loss is 1.   If a is chosen, the state of the world is not determined and is some number $0<s<1$.  If the state of the world is actually s, the loss is 2s. Following the $minmax$ principle, we get that l(a) = 2 and l(b) = 1 and it is rational for the decision maker to chose b.  If the decision maker assumes a uniform distribution over the possible values of s, then on the average the decision maker will loss 1 given she chosen a and she is indifference between the choices a and b. 
\end{example}

\begin{exercise}
In the previous example, assume that the decision maker is given the average and standard deviation of the distribution on as and models the distribution as the Beta distribution.  what will be the decision of the decision maker for various values of the average and standard deviation of the distribution s?
\end{exercise}

\begin{exercise}
Assume a binary classification and its associated confusion matrix. Can you define the Bayes rule associated with it and explain how it will be used to define the the usage of the classifier?
\end{exercise}

\href{https://www.youtube.com/watch?v=ztsGe4gOJJk&ab_channel=MLworkshop}{Click for YouTube recording on this topic}
\\

\section{Machine learning}

\subsection{Machine learning techniques}

\subsubsection{Hard support vector machine}

A binary classifier is obtain on data in $R^n$.  We first assume that there is a hyper plane that separates the training set to two different classes. This is referred to as hard support vector machine.  In the next subsection we relax that assumption.

More precisely given a training set $(x_1, y_1), \ldots, (x_m, y_m)$ that was sampled i.i.d. such that $x_i \in R^n$ and $y_i \in \{-1, 1\}$ we assume that there exist a hyper plane $(w, b), w \in R^n, b \in R$ such that $(w, x_i) + b > 0$  if and only if $y_i = 1$.   Here $(w, x_i)$ is the scalar multiplication of w and $x_i$.   

\begin{figure}[h!]
\centering
\includegraphics[scale=0.5]{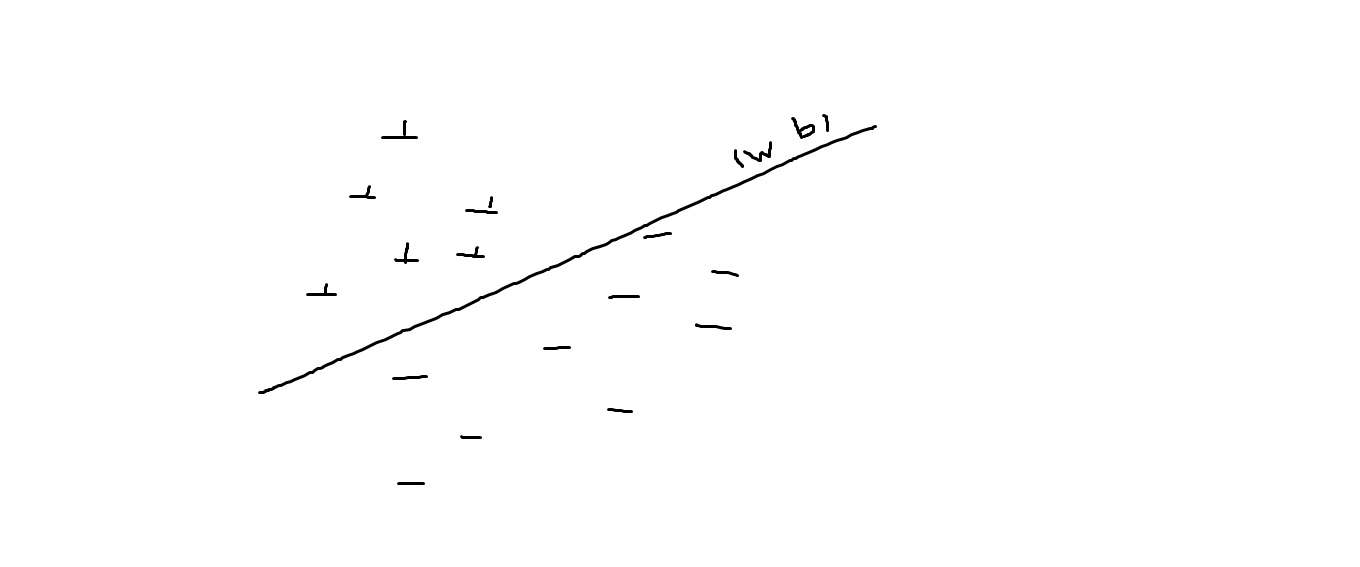}
\caption{A hyper plane exists that separates the positive and negative examples.}
\label{fig:hardSVM}
\end{figure}

Figure \ref{fig:hardSVM} represents the situation in $R^2$.  In the figure, the plus sign stands for the label 1 and the minus sign represents the label -1. 

A separating hyper plan is used as a classifier by calculating $(w x)+b$ on a new point $x$.   If the result is positive we say that the label 1 otherwise that the label is $-1$.

\begin{figure}[h!]
\centering
\includegraphics[scale=0.10]{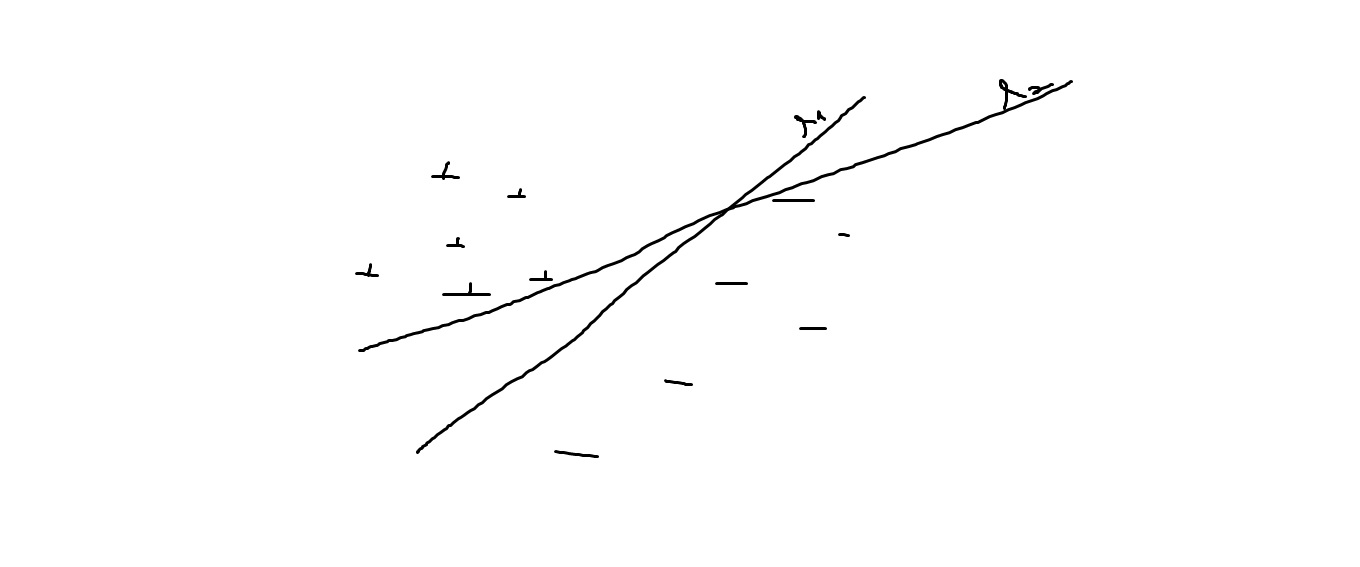}
\caption{$l_2$ is more sensitive to noise than $l_1$.}
\label{fig:hardSVMnoise}
\end{figure}

We would like to choose a separating hyper plane that will be as resilient as possible to noise.  Consider figure  \ref{fig:hardSVMnoise}.  Hyper plane $l_2$ is more sensitive to noise as it is closer to points in the data set.  If the points $l_2$ is close to change a bit due to noise, their classification by $l_2$ will change.  In contrast $l_2$ is more "in the middle" of the gap between the set that is positively labeled and the set that has negative label.   Thus, $l_1$ is less sensitive to noise. This intuition is captured in the following hard support vector machine optimization problem.

Denote the distance between the hyper lane $(w, b)$ and a point $x$ by $d((w, b), x)$.  We would like to find a hyper plane such that the closest point from a point in the training set to it will be as far away as possible.  In other words a hyper plane such that $max_{w \in R^n b \in R} min_{x_i, i = 1, \ldots, m} d((w, b), x_i)$.

\begin{exercise}
What is a hyper plan on the line?  Create a sample of points from the line as follows.  With equal probability randomly choose from normal distribution with average 0 and standard deviation 1 or from normal distribution with average 2 and standard deviation 1.  Obtain a sample of size 100.  Was the sample separable?  Repeat the process a few times.  Did the answer change?  Next change the average of the second distribution to 4, 8, 16, 32.   Did you get a separable sample in these cases?    
\end{exercise}

\begin{solution}
Solution to exercise

\definecolor{LightGray}{gray}{0.9}
\begin{minted}
[
frame=lines,
framesep=2mm,
baselinestretch=1.2,
bgcolor=LightGray,
fontsize=\footnotesize,
linenos
]
{python}

# Imports
%matplotlib inline
import numpy as np
import matplotlib.pyplot as plt

### The is the random function. 
# It generates n samples.
# Each samples has 50-50 probability to be taken from Distribution # 1 or Distribution 2.

def rand(n=10, avg_1=0, avg_2=2, std_1=1, std_2=1):
  r1 = np.random.normal(avg_1, std_1, size=n)  # N samples from dist. 1
  r2 = np.random.normal(avg_2, std_2, size=n)  # N samples from dist. 2
  c1 = np.random.choice([0, 1], size=n)  # Determine which samples will be taken from dist. 1
  c2 = 1 - c1  # Determine which samples will be taken from dist. 2
  r = (r1 * c1) + (r2 * c2)
  return r
  
### First experiment
# We randomize 100 samples from the function defined above and plot # the histogram we see. 
# 
# We repeat 5 times.

repeats = 5
samples_per = 1000
means = [0,2]
fig, axs = plt.subplots(1, repeats,figsize=(50,10))

for i in range(repeats):
  avg_2 = means[1]
  r = rand(n=samples_per,avg_2=avg_2)

  n_bins = 80
  ax = axs[i]
  # We can set the number of bins with the *bins* keyword argument.
  ax.set_title(f"(# {i+1})Histogram of points (mu1 = 0 , mu2 = {avg_2})")
  _ = ax.hist(r, bins=n_bins)
# See plots in the first figure below
  
### Second experiment
# We randomize 1000 samples from the function defined above and
# plot the histogram we see. 
# 
# We test what happens when the first distribution is always (0,1), but the second
# distribution can have a mean of [4,8,16,43].
# We repeat 4 times for each mean.
#
# We can see that the bigger the difference between the means,
# the easier it is to distinguish the two distributions.

repeats = 4
samples_per = 1000
second_means = [4,8,16,32]
fig, axs = plt.subplots(len(second_means), repeats,figsize=(50,8 * len(second_means)))

n_bins = 80
for j in range(len(second_means)):
  second_mean = second_means[j]
  for i in range(repeats):
    r = rand(n=samples_per,avg_2=second_mean)

    ax = axs[j,i]
    # We can set the number of bins with the *bins* keyword argument.
    ax.set_title(f"(# {i+1})Histogram of points (mu1 = 0 , mu2 = {second_mean})")
    _ = ax.hist(r, bins=n_bins)
    
# See plots in the second figure below

\end{minted} 
\end{solution}

\begin{figure}[h!]
\centering
\includegraphics[width=\textwidth]{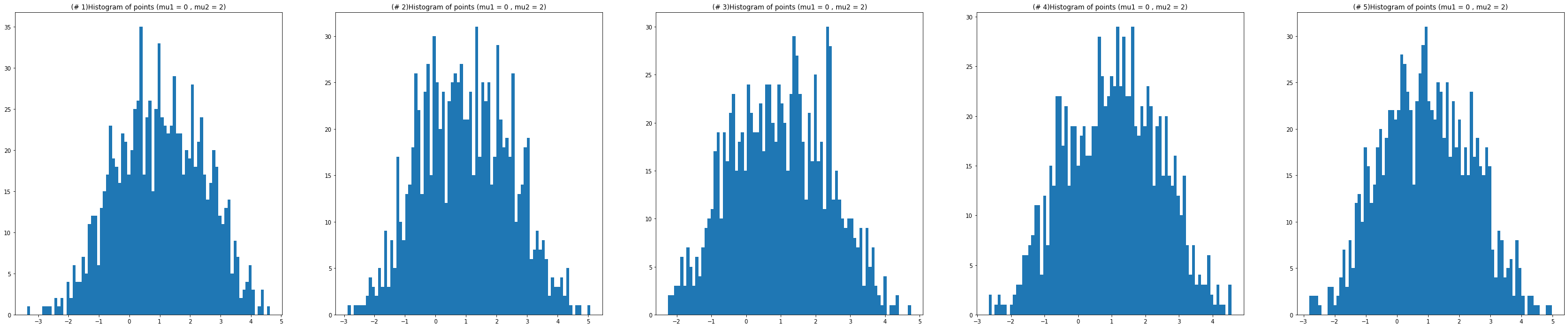}
\caption{5 repeats of joint distribution of two normal dist. with mu1=0,mu2=2}
\label{fig:code_a522_1}
\end{figure}

\begin{figure}[H]
\centering
\includegraphics[width=\textwidth]{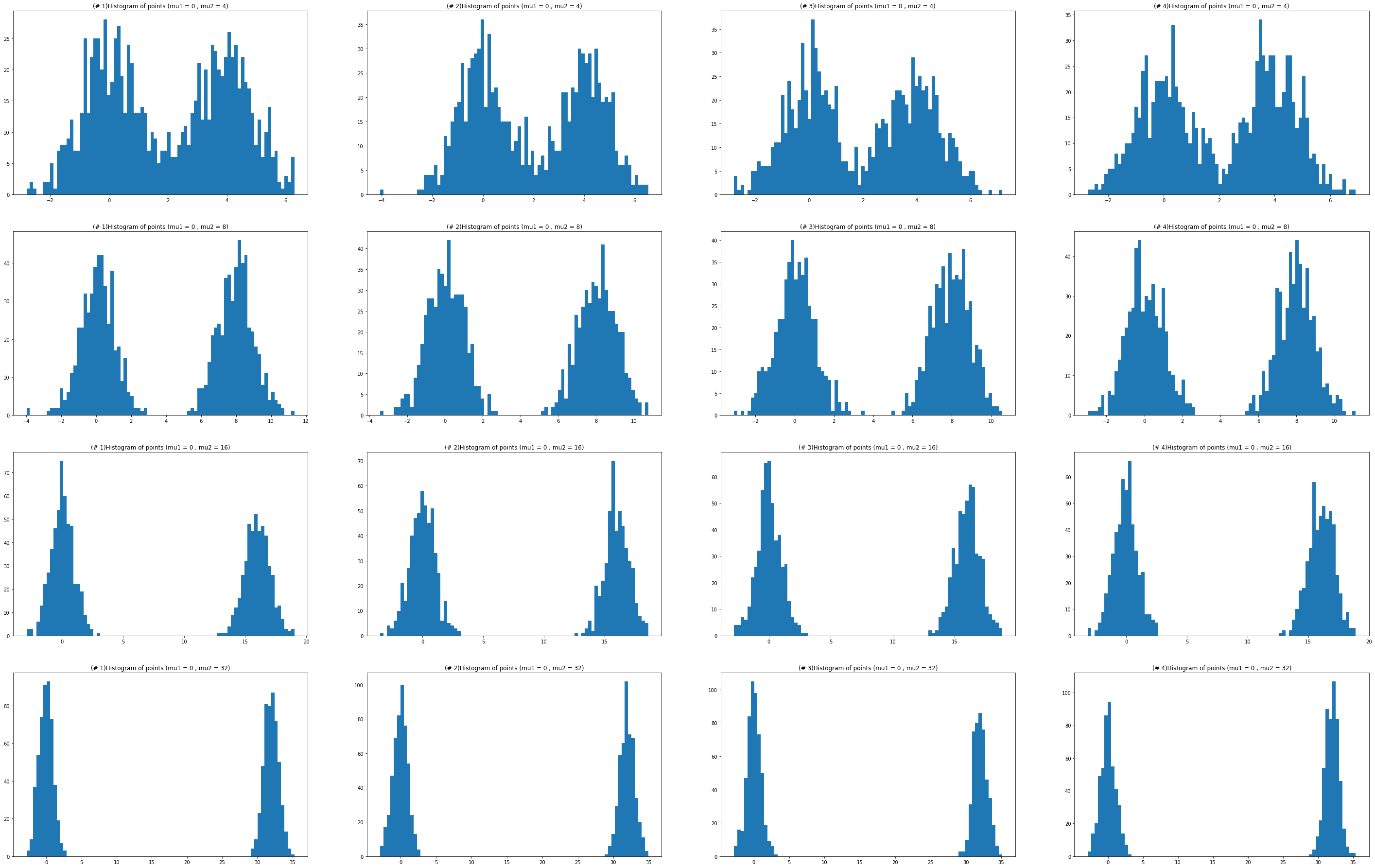}
\caption{Each row represents 4 repeats of the histogram of a joint distribution of two normal dist. with constant Mu. where mu1=0 in all of them, and in rows 1-4 the second mu is 4,8,16,32 respectively}
\label{fig:code_a522_2}
\end{figure}

\href{https://colab.research.google.com/drive/1bvXVLMZO27egVnzCN7oLpa36P6nyKevE?usp=sharing}{Click here for code}
\\

TBC - change to optimization problem due to the use a formula of the distance between a point and the plane

TBC - change to optimization problem to make it convex

TBC - Stress that this is an example of a ML related optimization problem.

\chapter{Solutions}
\label{chapter:solution_appendix}
Following are solutions to selected exercises.  

\begin{solution}
Solution to exercise \ref{Chapter:MLtesting:Excercise:distribution}.  $F(-1) = P(X \leq -1) = 0$ as the smallest possible value of X is 1.  Same goes for 0. In addition, $F(10.1) = P(X \leq 10.1) = P(\{1, 2, 3, 4, 5, 6\}) = \frac{6}{6} = 1$.  Same applies for 10, and 6.  Now $F(2) = P(X \leq 2) = P(\{1, 2\}) = \frac{2}{6}$.  Similarly, $F(1) = \frac{1}{6}$, $F(5) = P(X \leq 5) = P(\{1, 2, 3, 4, 5\}) = \frac{5}{6}$. 
\end{solution}

\begin{solution}
Solution to exercise \ref{mlTesting:dataBalance}.  
\begin{enumerate}
    \item
The probability of getting a cat image in one sampling is $\frac{200}{200+800} = \frac{1}{5}$.  Consider the indicator variable $X_i$ which is one if we got a cat image in the i sample and 0 otherwise.  We get the $E(X_i) = 1 \times \frac{1}{5} + 0 \times \frac{4}{5} = \frac{1}{5}$. The average number of cat images in 2000 samples with replacements is $E( \frac {\sum^{i = 2000}_{i = 1} X_i}{2000}) $.  From the linearity of expectation we get that this is equal to $\frac{\sum^{i = 2000}_{i = 1} E(X_i)}{2000}$ which is equal to $\frac{2000 \times E(X_i)}{2000}$.  This is then equal to $\frac{1}{5}$.  
    \item 
We first decide if to sample a cat image or a dog image randomly (with equal probabilities).  If in the first step we decided to sample a cat image we sample with replacement from the 200 cat images, otherwise we sample with replacement from the 800 dog images.  The average of $X_i$ is now clearly $\frac{1}{2}$.  Following the same procedure as in the previous part of the question we get that the overall average is $\frac{1}{2}$.
\item This procedure is used to create a training set that have a balanced number of cat images and dog images. 
\end{enumerate}
\end{solution}

\clearpage
\begin{solution}
\label{estimatePrecentage_sol}
Solution to exercise \ref{estimatePrecentage}

\definecolor{LightGray}{gray}{0.9}
\begin{minted}
[
frame=lines,
framesep=2mm,
baselinestretch=1.2,
bgcolor=LightGray,
fontsize=\footnotesize,
linenos
]
{python}

# Imports
import numpy as np

# Given a number, x, a function returns x^2 most of the time,
# i.e., in p percentage of the cases.  Write a test that finds the percentage p.
def foo(x):
    p = 0.40
    if np.random.uniform(0,1) < p:
        return x ** 2
    else:
        return -1
        
iterations = 10000
hits = 0
for i in range(iterations):
    if foo(3) == 9:
        # Notice that any number could have been used instead of 3
        # 3 is an example and can be replaced.
        hits += 1
    else:
        pass

p_hat = hits / iterations
print(f"Estimation for p: {p_hat:>.3f}")
\end{minted} 

Output:
\begin{minted}
[
frame=lines,
framesep=2mm,
baselinestretch=1.2,
bgcolor=LightGray,
fontsize=\footnotesize,
linenos
]
{python}
Estimation for p: 0.390
\end{minted} 

\textbf{Question}
 How is that different from a test that checks that a deterministic function correctly evaluates $x^2$ ?
 
\textbf{Answer}
This function might give a different answer for the SAME results, and therefore repeating the same input is meaningful.
\end{solution}
\clearpage
\clearpage
\begin{solution}
\label{Chapter:MLtesting:Excercise:parabola_sol}
Solution to exercise \ref{Chapter:MLtesting:Excercise:parabola}

\definecolor{LightGray}{gray}{0.9}
\begin{minted}
[
frame=lines,
framesep=2mm,
baselinestretch=1.2,
bgcolor=LightGray,
fontsize=\footnotesize,
linenos
]
{python}

# Imports
%matplotlib inline
import numpy as np
import matplotlib.pyplot as plt
from tqdm import tqdm

# The function f(x) = a(x^2) + b*x + c
a = 1
b = -2
c = -3

# The range examined
x_range = -5, 5

# Function f implementation
def f(x):
    v = (a * (x ** 2)) + (b * x) + c
    return v
    
###  Parabola visualization 
Xi = np.arange(x_range[0], x_range[1], 0.01)
y = np.array(list(map(f, Xi)))
plt.title(f'f(x) = {a:+}(x^2) {b:+}x {c:+}')
plt.plot(Xi, y)
plt.grid()
plt.show() # See img 1 in the figure below

# Now we fit a Linear regression
from sklearn.metrics import mean_squared_error

# Generate D (the data, marked as X)
D_size = 100  # Number of samples
X = np.random.uniform(x_range[0], x_range[1], D_size)
y = list(map(f, X))

# Define g() as the function of the linear regression
d, e = np.polyfit(X, y, 1)

def g(x):
    return (d * x) + e

print(f"Model: g(x) = {d:>.3f}x + {e:>.3f}")
# Output: 
# Model: g(x) = -1.673x + 4.439

yhat = np.array(list(map(g, X)))
y = np.array(list(map(f, X)))
train_mse = mean_squared_error(y, yhat)
print(f"Train Mean Square Error: {train_mse:>.3f}")
# Output: 
# Train Mean Square Error: 41.316

### Visualizing just g()
yhat = np.array(list(map(g, Xi)))
plt.title(f'g(x) = {d:>.3f}x + {e:>.3f}')
plt.plot(Xi, yhat, color='r')
plt.grid()
plt.show() # See img 2 in the figure below

yhat = np.array(list(map(g, Xi)))
y = np.array(list(map(f, Xi)))

### Visualize the overlap between f() and g()
plt.title(f'Mean Square Error on train: {train_mse:>.3f}')
plt.plot(Xi, yhat, color='r', label='Regression line')
plt.plot(Xi, y, color='b', label='parabola')
plt.grid()
plt.legend()
plt.show() # See img 3 in the figure below

### Re-define g as a function of x,x^2
# g'(x)=g(x,x2)=a*(x2)+b*(x)+c

# Generate D (the data, marked as X)
D_size = 5
Xi = np.random.uniform(x_range[0], x_range[1], D_size)
X = np.zeros((D_size, 2))
X[:, 0] = Xi
X[:, 1] = Xi ** 2
y = np.array(list(map(f, Xi)))

# Train a linear regression of 2 parameters
from sklearn import linear_model

regr = linear_model.LinearRegression()
_  = regr.fit(X, y)

bg, ag = regr.coef_
cg = regr.intercept_

def g(x):
    return (ag * (x ** 2) + bg * x + cg)

Xi = np.arange(x_range[0], x_range[1], 0.01)
yhat = np.array(list(map(g, Xi)))
y = np.array(list(map(f, Xi)))

train_mse = mean_squared_error(y, yhat)
print(f"Train Mean Square Error: {train_mse:>.3f}")
# Output: 
# Train Mean Square Error: 0.000

### Visualize the new g()
yhat = np.array(list(map(g, Xi)))
plt.title(f'g(x) = ({ag:+>.3f})x^2 + ({bg:+>.3f})x + ({cg:>+.3f})')
plt.plot(Xi, yhat, color='r')
plt.grid()
plt.show() # See img 4 in the figure below

# Overlap f() and the new g()
plt.close('all')
plt.title(f'Mean Square Error on train: {train_mse:>.3f}')
plt.plot(Xi, yhat, color='r', label='Regression line')
plt.plot(Xi, y, color='b', label='parabola')
plt.grid()
plt.legend()
plt.show() # See img 5 in the figure below

### Now, what will happen if we add some noise?
# Let's define the new and noisy f function

def f(x,noise_mu=0,noise_sigma=1):
    noise = np.random.normal(loc=noise_mu,scale=noise_sigma)
    v = (a * (x ** 2)) + (b * x) + c + noise
    return v
    
# plot the noisy image
Xi = np.arange(x_range[0], x_range[1], 0.01)
y = np.array(list(map(f, Xi)))
plt.title(f'f(x) = {a:+}(x^2) {b:+}x {c:+}')
plt.plot(Xi, y)
plt.grid()
plt.show() # See img 6 in the figure below

# Generate D (the data, marked as X)
D_size = 250
Xi = np.random.uniform(x_range[0], x_range[1], D_size)
X = np.zeros((D_size, 2))
X[:, 0] = Xi
X[:, 1] = Xi ** 2
y = np.array(list(map(f, Xi)))

# Create a new g'(x) = g(x,x^2) = a *(x^2) + b*(x) + c.
# Except, now the y label is noisy because of the new f function
regr = linear_model.LinearRegression()
_  = regr.fit(X, y)

bg, ag = regr.coef_
cg = regr.intercept_

def g(x):
    return (ag * (x ** 2) + bg * x + cg)
    
# Measure the MSE
Xi = np.arange(x_range[0], x_range[1], 0.01)
yhat = np.array(list(map(g, Xi)))
y = np.array(list(map(f, Xi)))

train_mse = mean_squared_error(y, yhat)
print(f"Train Mean Square Error: {train_mse:>.3f}")
# Output: 
# Train Mean Square Error: 0.916

plt.close('all')
# plt.title(f'Mean Square Error on train: {train_mse:>.3f}')
plt.title(f'g(x) = ({ag:+>.3f})x^2 + ({bg:+>.3f})x + ({cg:>+.3f})')
plt.plot(Xi, y, color='b', label='Noisy parabola')
plt.plot(Xi, yhat, color='r', label='Regression line',linewidth=3.0)
plt.grid()
plt.legend()
plt.show() # See img 7 in the figure below

# The resulting regression model is a close match. But the noise causes slight error. 
# If we were to increase the size of D, the dataset - the MSE will decrease as well

### Comparing the size of the noisy dataset to the size of the error (MSE)
sizes = 100
bar = range(2,sizes)
measures = [0] * len(bar)
d_sizes = [0] * len(bar)

for idx, D_size in tqdm(enumerate(bar),total=len(bar)):
  repeat = 50
  reps = [0] * repeat
  for repeat_idx in range(repeat):
    # Generate D (the data, marked as X)
    Xi = np.random.uniform(x_range[0], x_range[1], D_size)
    X = np.zeros((D_size, 2))
    X[:, 0] = Xi
    X[:, 1] = Xi ** 2
    y = np.array(list(map(f, Xi)))

    regr = linear_model.LinearRegression()
    _  = regr.fit(X, y)
    bg, ag = regr.coef_
    cg = regr.intercept_

    def g(x):
        return (ag * (x ** 2) + bg * x + cg)

    Xi = np.arange(x_range[0], x_range[1], 0.01)
    yhat = np.array(list(map(g, Xi)))
    y = np.array(list(map(f, Xi)))

    train_mse = mean_squared_error(y, yhat)
    reps[repeat_idx] = train_mse
  train_mse = np.mean(reps)
  d_sizes[idx] = D_size
  measures[idx] = train_mse
 
plt.close('all')
plt.title(f'Average MSE compared to dataset size')
plt.plot(d_sizes, measures, color='b', label='MSE')
plt.xlabel('Size of dataset')
plt.ylabel('Averge MSE')
plt.ylim(0,10)
plt.grid()
plt.legend()
plt.show() See img 8 in the figure below
  
\end{minted} 
\end{solution}

\begin{figure}[H]
  \begin{subfigure}[b]{0.4\textwidth}
    \includegraphics[width=\textwidth]{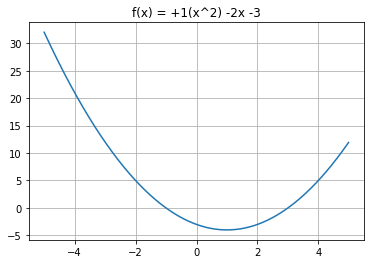}
    \caption{code plot 1}
    \label{fig:ex_232_1}
  \end{subfigure}
  \hfill
  \begin{subfigure}[b]{0.4\textwidth}
    \includegraphics[width=\textwidth]{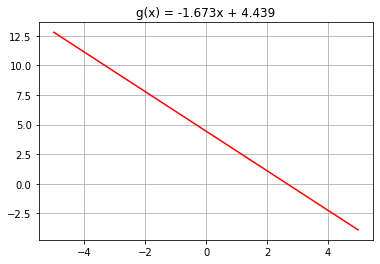}
    \caption{code plot 2}
    \label{fig:ex_232_2}
  \end{subfigure}
  
  \begin{subfigure}[b]{0.4\textwidth}
    \includegraphics[width=\textwidth]{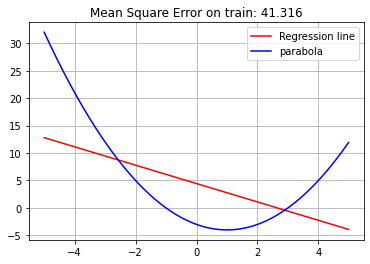}
    \caption{code plot 3}
    \label{fig:ex_232_3}
  \end{subfigure}
  \hfill
  \begin{subfigure}[b]{0.4\textwidth}
    \includegraphics[width=\textwidth]{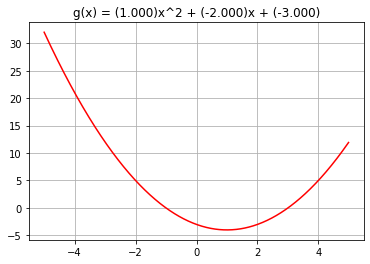}
    \caption{code plot 4}
    \label{fig:ex_232_4}
  \end{subfigure}
  
  \begin{subfigure}[b]{0.4\textwidth}
    \includegraphics[width=\textwidth]{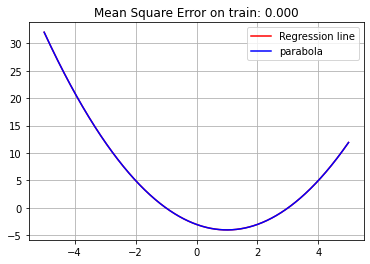}
    \caption{code plot 5}
    \label{fig:ex_232_5}
  \end{subfigure}
  \hfill
  \begin{subfigure}[b]{0.4\textwidth}
    \includegraphics[width=\textwidth]{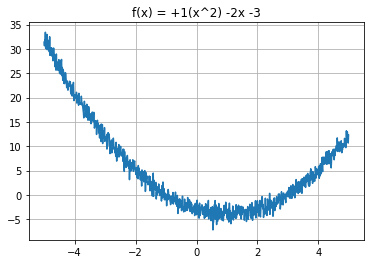}
    \caption{code plot 6}
    \label{fig:ex_232_6}
  \end{subfigure}
  
  \begin{subfigure}[b]{0.4\textwidth}
    \includegraphics[width=\textwidth]{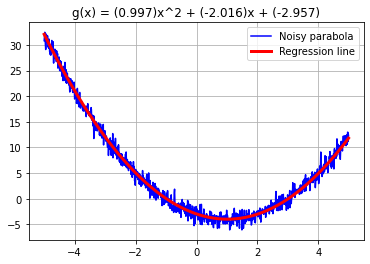}
    \caption{code plot 7}
    \label{fig:ex_232_7}
  \end{subfigure}
  \hfill
  \begin{subfigure}[b]{0.4\textwidth}
    \includegraphics[width=\textwidth]{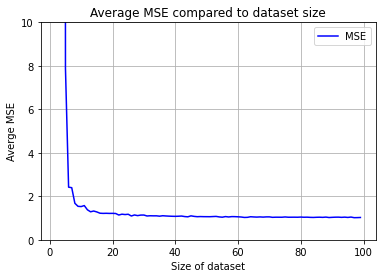}
    \caption{code plot 8}
    \label{fig:ex_232_8}
  \end{subfigure}
  
  \caption{Code image plots}
\end{figure}

\clearpage
\begin{solution}
\label{excercise:402_sol}
Solution to exercise \ref{excercise:402}

\definecolor{LightGray}{gray}{0.9}
\begin{minted}
[
frame=lines,
framesep=2mm,
baselinestretch=1.2,
bgcolor=LightGray,
fontsize=\footnotesize,
linenos
]
{python}

# Imports
import numpy as np
import pandas as pd
from scipy.optimize import minimize

%matplotlib inline
import matplotlib.pyplot as plt

### Define the function to predict
class F_class:
    def __init__(self, a, A, B):
        self.a = a
        self.A = A
        self.B = B

    def calc(self, x):
        res = np.zeros(x.shape)

        bigger_idx = x > self.a
        smaller_idx = x <= self.a

        res[bigger_idx] = np.power(x[bigger_idx], 2) * self.A
        res[smaller_idx] = np.power(x[smaller_idx], 2) * self.B

        return res
### The nature-given parameters
a = 0
A = 1
B = -1

# The dataset D
f = F_class(a, A, B) # The function with given parameters
X = np.arange(-5, 5)
Y = f.calc(X)
D = pd.DataFrame(columns=['X', 'Y'], data=np.array([X, Y]).T) # See dataset in img (a) below

### The ML
def sum_of_squares(params, X, Y):
  a, A, B = params
  model = F_class(a, A, B)
  y_pred = model.calc(X)
  obj = np.sqrt(((y_pred - Y) ** 2).sum())
  return obj

# perform fit to find optimal parameters
# initial value is a guess
initial_guess = [0., 0., 0.]  # a, A, B
res = minimize(sum_of_squares, x0=initial_guess, args=(X, Y), tol=1e-5, method="Powell")

### ML evaluation
a_pred, A_pred, B_pred = res.x
model = F_class(a_pred, A_pred, B_pred)
Y_pred = model.calc(X)
MSE = np.sqrt(((Y_pred - Y) ** 2).sum())

print("Estimated values:")
print(f"a = {a_pred:>.3f}")
print(f"A = {A_pred:>.3f}")
print(f"B = {B_pred:>.3f}")
print(f"MSE: {MSE}")

### Print output:
# Estimated values:
# a = 0.990
# A = 1.000
# B = -1.000
# MSE: 7.472741870478288e-11

plt.plot(X, Y)
plt.plot(X, Y_pred,  'bo')
plt.ylim(min(Y), max(Y))
plt.xlabel('x')
plt.ylabel('Function estimation')
plt.axhline(y=a, color='r', linestyle='-')
plt.title('f(x)')
plt.show() # See plot in img (b) below
\end{minted} 
\end{solution}

\begin{figure}[H]
  \begin{subfigure}[b]{0.4\textwidth}
    \includegraphics[width=20mm,scale=0.05]{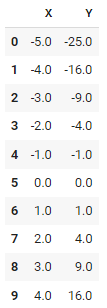}
    \caption{Dataset D}
    \label{fig:ex_402_1}
  \end{subfigure}
  \hfill
  \begin{subfigure}[b]{0.4\textwidth}
    \includegraphics[width=\textwidth]{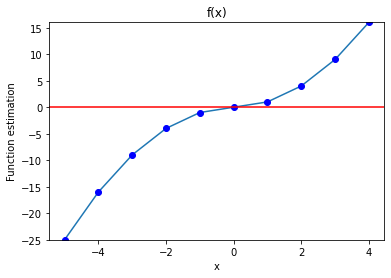}
    \caption{Etimation of f}
    \label{fig:ex_402_2}
  \end{subfigure}
  \caption{Code image plots}
\end{figure}

\clearpage
\begin{solution}
\label{Chapter:MLtesting:Excercise:emericalDistribution_sol}
Solution to exercise
\ref{Chapter:MLtesting:Excercise:emericalDistribution} can be found
 \href{https://colab.research.google.com/drive/1Izd85vZ6qW77V6z8auEbC5yfPm6ldJxO?usp=sharing}{\textbf{Here}} or in the code below.
\end{solution}

\definecolor{LightGray}{gray}{0.9}
\begin{minted}
[
frame=lines,
framesep=2mm,
baselinestretch=1.2,
bgcolor=LightGray,
fontsize=\footnotesize,
linenos
]
{python}

# Imports
%matplotlib inline
import numpy as np
import matplotlib.pyplot as plt

# Code support functions 
def dice(n=1):
    # A fair dice. n is the number of throws
    return np.random.choice([1, 2, 3, 4, 5, 6], n)


def is_even(n=1):
    # return 1 if it gets an even number in a fair dice toss, 0 otherwise
    # n is the number of throws
    return 1 - (dice(n) % 2)


def is_odd(n=1):
    # return 1 if it gets an even number in a fair dice toss, 0 otherwise
    # n is the number of throws
    return (dice(n) % 2)

# Code functions 
def F(x,S):
  return np.mean(np.array(S <= x,dtype=int))

def F0(S):
  return F(0,S)

def F1(S):
  return F(1,S)

# Define sample size
sample_size = 1000

# Estimate even/odd balance over N tosses
# with N in \{1,...,500\}
vector_length = list(range(1, sample_size))
p_hats = [F0(is_odd(nt)) for nt in vector_length]

# Plotting
plt.plot(vector_length, p_hats)
plt.ylim(0, 1)
plt.xlabel('Dice toss repeats')
plt.ylabel('Mean value of "is odd on dice"')
plt.axhline(y=0.5, color='r', linestyle='-')
plt.title('Fe(0)')
plt.show() # Results available below

# Now for Fe(1) - Fe(0)
vector_length = list(range(1, sample_size))
p_hats = [F1(is_odd(nt)) - F0(is_odd(nt)) for nt in vector_length]

# Plotting
plt.plot(vector_length, p_hats)
plt.ylim(0, 1)
plt.xlabel('Dice toss repeats')
plt.ylabel('F(1) - F(0)')
plt.axhline(y=0.5, color='r', linestyle='-')
plt.title('Fe(1) - Fe(0)')
plt.show() # Results available below
\end{minted} 

\begin{figure}[h!]
  \begin{subfigure}[b]{0.4\textwidth}
    \includegraphics[width=\textwidth]{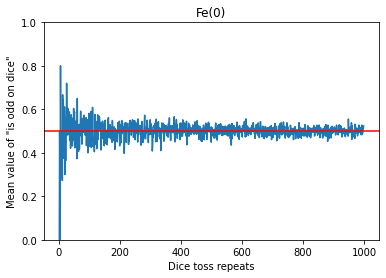}
    \caption{Code output for $Fe(0)$}
    \label{fig:622_f0}
  \end{subfigure}
  \hfill
  \begin{subfigure}[b]{0.4\textwidth}
    \includegraphics[width=\textwidth]{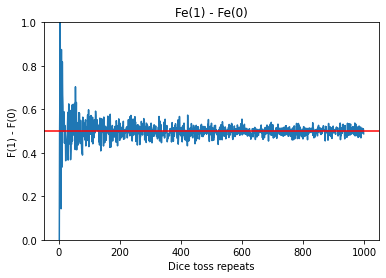}
    \caption{Code output for $Fe(0) - Fe(0)$}
    \label{fig:622_f1_minus_f0}
  \end{subfigure}
  \caption{Code image plots}
\end{figure}
\clearpage
\begin{solution}
\label{excercise:525_gn_fn_sol}
Solution to exercise \ref{excercise:525_gn_fn}

\definecolor{LightGray}{gray}{0.9}
\begin{minted}
[
frame=lines,
framesep=2mm,
baselinestretch=1.2,
bgcolor=LightGray,
fontsize=\footnotesize,
linenos
]
{python}

# Imports
%matplotlib inline
import matplotlib.pyplot as plt
import numpy as np

### The series of function fn(x)
class Fn:
  def __init__(self,n):
    self.n=float(n)

  def calc(self,x):
    x = float(x)
    return ((0.5)*((1.0 - (1.0/self.n)) ** 2) * x) + ((1.0 - (1.0/self.n)))
    
### The series of function gn(x)
class Gn:
  def __init__(self,n):
    self.n=float(n)

  def calc(self,x):
    x = float(x)
    lower_threshold = - 1.0 / (0.5*(1.0 - (1.0/n)))
    upper_threshold = 0
    if (lower_threshold <= x) and (x <= upper_threshold):
      return ((0.5)*((1.0 - (1.0/self.n)) ** 2) * x) + ((1.0 - (1.0/self.n)))
    else:
      return 0
      
### The function f(x)
def f(x):
  x = float(x)
  return (0.5*x) + 1
  
### The function g(x)
def g(x):
  x = float(x)
  lower_threshold = -2
  upper_threshold = 0
  if (lower_threshold <= x) and (x <= upper_threshold):
    return f(x)
  else:
    return 0

### The limit of fn(x)
n_bound = 50

# limit function
x1 = -2
x2 = 0
y1 = f(x1)
y2 = f(x2)
plt.plot([x1,x2],[y1, y2],'b--',label='limit function, f(x)')

for n in range(2,n_bound+1):
  x1 = -2
  x2 = 0

  fn = Fn(n)
  y1 = fn.calc(x1)
  y2 = fn.calc(x2)
  
  plt.plot([x1,x2],[y1, y2])
plt.ylabel('some numbers')
plt.legend()
plt.show() # See img (a) below

### g(x) and f(x) compared
n_bound = 50

# limit function
x = np.arange(-5,5)
fy = [f(xt) for xt in x]
gy = [g(xt) for xt in x]
plt.plot(x,fy,label='f(x)')
plt.plot(x,gy,label='g(x)')

plt.ylabel('some numbers')
plt.legend()
plt.show() # See img (b) below

### The limit of gn(x)
n_bound = 50

# limit function
x = np.arange(-5,5)
gy = [g(xt) for xt in x]
plt.plot(x,gy,'b--',label='limit function, g(x)',linewidth=3.5)

for n in range(2,n_bound+1):
  gn = Gn(n)
  gny = [gn.calc(xt) for xt in x]
  
  plt.plot(x,gny)
plt.ylabel('some numbers')
plt.legend()
plt.show() # See img (c) below
\end{minted} 
\end{solution}

\begin{figure}[H]
  \begin{subfigure}[b]{0.4\textwidth}
    \includegraphics[width=\textwidth]{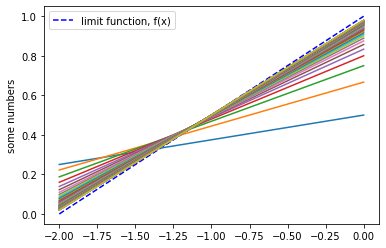}
    \caption{The limit of fn(x)}
    \label{fig:ex_524_1}
  \end{subfigure}
  \hfill
  \begin{subfigure}[b]{0.4\textwidth}
    \includegraphics[width=\textwidth]{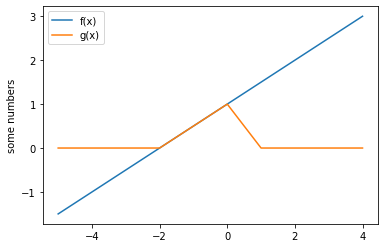}
    \caption{gn(x) and fn(x)}
    \label{fig:ex_524_2}
  \end{subfigure}
  
  \begin{subfigure}[b]{0.4\textwidth}
    \includegraphics[width=\textwidth]{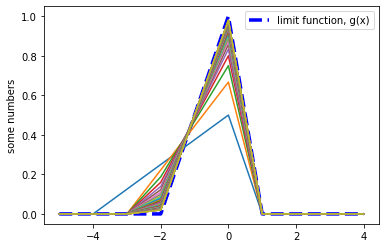}
    \caption{The limit of gn(x)}
    \label{fig:ex_524_3}
  \end{subfigure}
  \hfill
  
  \caption{Code image plots}
\end{figure}

\clearpage
\begin{solution}
\label{excercise:622_sol}
Solution to exercise \ref{excercise:622}
\definecolor{LightGray}{gray}{0.9}
\begin{minted}
[
frame=lines,
framesep=2mm,
baselinestretch=1.2,
bgcolor=LightGray,
fontsize=\footnotesize,
linenos
]
{python}

# Imports
%matplotlib inline

import numpy as np
import pandas as pd
import matplotlib.pyplot as plt

# Parameters
x_mean = 0
x_variance = 1  # This is epsilon, set to epsilon = 1

t_mean = 0
t_variance = 100

z_mean = 0
z_variance = 1000

n = 10000  # Number of samples

# Generate databases
X = np.random.normal(x_mean, x_variance, n)
T = np.random.normal(t_mean, t_variance, n)
Z = np.random.normal(z_mean, z_variance, n)
df = pd.DataFrame(columns=['X', 'T', 'Z'], index=range(n), data=np.array([X, T, Z]).T)

# Plot X, T and Z distribution
n_bins = 20
fig, axs = plt.subplots(1, 1, sharey=True, tight_layout=True)

axs.hist(Z, bins=n_bins,label='Z')
axs.hist(T, bins=n_bins,label='T')
axs.hist(X, bins=n_bins,label='X')

plt.title("X, T and Z histogram compared")
plt.legend()
plt.show() # See img (a) in the figure below

# Create Y = X + T + Z
df['Y'] = df.sum(axis=1)

# Create Y' = Average(X) + T + Z
X_avg = df['X'].sum() / n
df['Y\''] = X_avg + df[['T', 'Z']].sum(axis=1)

# Compare Y and Y'
Y_diff = df['Y'] - df['Y\'']
 
# Plot the difference between Y and Y' 
n_bins = 25
fig, axs = plt.subplots(1, 1, sharey=True, tight_layout=False)

mx = Y_diff.max()
min = Y_diff.min()
mean = Y_diff.mean()
std = Y_diff.std()

msg = f'Y - Y\'' + '\n'
msg += f'Max diff:       {mx:>.2f}' + '\n'
msg += f'Min diff:       {min:>.2f}' + '\n'
msg += f'Mean of diff: {mean:>.2f}' + '\n'
msg += f'Std of diff:    {std:>.2f}' + '\n'

axs.hist(Y_diff, bins=n_bins)
axs.text(5,int(n/15),msg,bbox={'facecolor': 'red', 'alpha': 0.5, 'pad': 10},size=15)

plt.title("Distribution of Y - Y\'")
plt.show() # See img (b) in the figure below

\end{minted}

\textbf{Question}
What is the maximal
difference between $yi$ and $y'$
i? what does it tell you about the learning of y?
 
\textbf{Answer}
The maximal difference is $~4$. What is interesting is that the learned $mean$ of $y $ and $y'$ is the same. But the $Std$ of of the difference is equal to the $Std$ that was removed - that of $X$

\begin{figure}[H]
  \begin{subfigure}[b]{0.4\textwidth}
    \includegraphics[width=\textwidth]{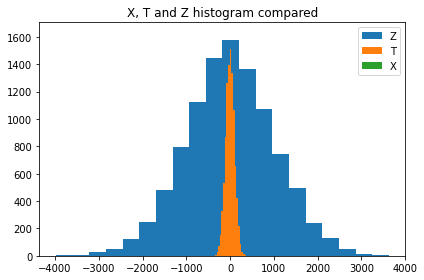}
    \label{fig:ex_622_1}
  \end{subfigure}
  \hfill
  \begin{subfigure}[b]{0.4\textwidth}
    \includegraphics[width=\textwidth]{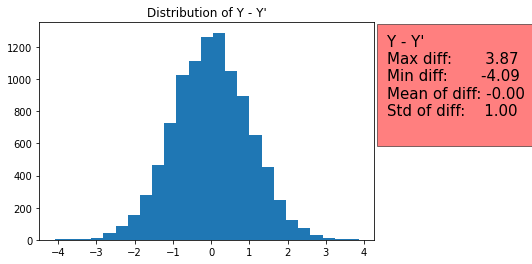}
    \label{fig:ex_622_2}
  \end{subfigure}
  
  \caption{Code image plots}
\end{figure}

\end{solution}

\clearpage
\begin{solution}
\label{excercise:a22_sol}
Solution to exercise \ref{excercise:a22}
\definecolor{LightGray}{gray}{0.9}
\begin{minted}
[
frame=lines,
framesep=2mm,
baselinestretch=1.2,
bgcolor=LightGray,
fontsize=\footnotesize,
linenos
]
{python}

# Imports
%matplotlib inline
import numpy as np
from scipy.special import comb
import matplotlib.pyplot as plt

### Parameters, can be replaced by other values
p = 0.01  # Probability of getting a defect
n = 1000  # products per batch in production line

alpha = 0.01  # likelihood of finding these or more defects

def prob_of_these_defects(t):
    # a function that calculated the probability of seeing exactly t defect in a production line
    coef_a = comb(n, t)
    coef_b = np.power(p, t)
    coef_c = np.power(1.0 - p, n - t)
    return coef_a * coef_b * coef_c


def prob_of_seeing_more_defects(k):
    # a function that calculated the prob of k or more defects
    t_vector = [prob_of_these_defects(t) for t in range(k, n + 1)]
    t_sum = np.sum(t_vector)
    return t_sum

### Run calculations
k_values = list()
k_results = list()
for k in range(2, int(n / 2)):
  likelihood_of_k_defects = prob_of_seeing_more_defects(k)
  k_results.append(likelihood_of_k_defects)
  k_values.append(k)

  if likelihood_of_k_defects < alpha * 0.001:
      break  # To save computational effort, when the likelihood drops significantly below alpha, we break
      
### Find optimal k
idx_of_closest = (np.abs(np.asarray(k_results) - alpha)).argmin()
optimal_k = k_values[idx_of_closest]
optimal_value = k_results[idx_of_closest]
# The value of the optimal k in this case is calculated to be 19

# plot
plt.plot(k_values, k_results, 'black', label='k-to-alpha', linewidth=3.5)
plt.axhline(y=alpha, color='r', label=f'alpha ({alpha:>.3f})')
plt.axvline(x=optimal_k, color='g', label=f'Optimal k ({optimal_k})')

plt.title(f'optimal K for desired alpha (n={n},p={p:>.3f})')
plt.ylabel('alpha')
plt.xlabel('k')
plt.legend()
plt.show() # See img in the figure below
\end{minted}

\textbf{Question}
Is there a computational limitation to the calculation
of k when n is large ?
 
\textbf{Answer}
Yes. As the calculations include factorial operations, when $n$ increase the computational increases dramatically. In the code above an average run with $n=1,000$ will usually take less than 1 second. But increasing this number to $n=10,000$ will take about 30 minutes. 

\begin{figure}[H]
  \begin{subfigure}[b]{0.6\textwidth}
    \includegraphics[width=\textwidth]{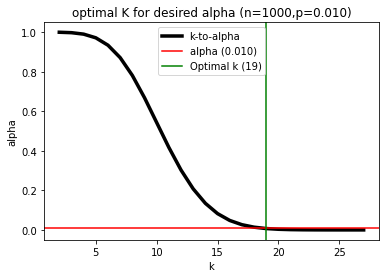}
    \caption{optimal $k$ for each $\alpha$}
    \label{fig:ex_a22_1}
  \end{subfigure}
  \caption{Code image plots}
\end{figure}

\end{solution}

\clearpage
\chapter{Videos}

In this chapter we provide videos that supports the text.  The videos are associated with the text through appropriate internal links.

The following videos discusses the testing of ML systems and is associated with chapter \ref{Chapter:MLtesting} introduction and section \ref{randomVariable}.

This video gives an overview of the chapter and discusses the concept of a random variable and its distribution. See link \href{https://www.youtube.com/watch?v=tW5_kKQHgtg&list=PLC7m9qp0Q1Ydn7ZtR9yeI46yHw-4q11iI&index=2}{here}.

This video covers basic properties of the expectation and variance of a random variable.  See link \href{https://www.youtube.com/watch?v=suGPMSPad70&list=PLC7m9qp0Q1Ydn7ZtR9yeI46yHw-4q11iI&index=6}{here}.
\chapter{Annotated References}

Here we briefly discuss some references on the crafting of reliable ML embedded solutions.  

In \cite{FRB2021} a general motivation for assurance of ML embedded systems is given.  

In \cite{DBLP:journals/corr/abs-2108-00941} a human in the loop in machine learning tutorial is given.  The impact of the economy of the organization is not discussed.

\printglossaries
\printbibliography
\end{document}